\documentclass[review,3p,authoryear]{elsarticle}

\makeatletter
\def\ps@pprintTitle{%
	\let\@oddhead\@empty
	\let\@evenhead\@empty
	\let\@oddfoot\@empty
	\let\@evenfoot\@empty
}

\usepackage{amssymb}
\usepackage{amsmath}

\usepackage[utf8]{inputenc} 
\usepackage[T1]{fontenc}    
\usepackage{hyperref}       
\usepackage{url}            
\usepackage{booktabs}       
\usepackage{makecell}
\usepackage{multirow}
\usepackage{amsfonts}       
\usepackage{nicefrac}       
\usepackage{microtype}      
\usepackage{xcolor}         
\usepackage{dsfont}
\usepackage{enumitem}
\setlist[itemize]{noitemsep, topsep=0pt, parsep=0pt, partopsep=0pt}

\usepackage{comment}

\usepackage{hyperref}
\usepackage{url}
\usepackage{xurl}
\usepackage{amsmath, amsthm, bbm, algorithm, algpseudocode}

\usepackage{amsmath, amssymb, amsthm}
\usepackage{mathtools}
\newtheorem{assumption}{Assumption}
\newtheorem{lemma}{Lemma}
\newtheorem{proposition}{Proposition}
\newtheorem{theorem}{Theorem}

\newcommand{\bx}{\mathbf{x}}

\newcommand{\bb}{\mathbf{b}}

\newcommand{\bbeta}{\boldsymbol{\beta}}

\begin{document}

\begin{frontmatter}



\title{E-QRGMM: Efficient Generative Metamodeling for Covariate-Dependent Uncertainty Quantification} 


\author[label1]{Zhiyang Liang} 
\ead{liangzy23@m.fudan.edu.cn}
\author[label2,label3,label4]{Qingkai Zhang\corref{cor1}} 
\ead{22110690021@m.fudan.edu.cn}
\cortext[cor1]{Corresponding author}
\affiliation[label1]{organization={School of Data Science},
            addressline={Fudan University}, 
            postcode={Shanghai 200433}, 
            country={China}}

\affiliation[label2]{organization={School of Management},
            addressline={Fudan University}, 
            postcode={Shanghai 200433}, 
            country={China}}
            
\affiliation[label3]{organization={Department of Decision Analytics and Operations, College of Business, City University of Hong Kong},
             city={Hong Kong},
             country={China}}
\affiliation[label4]{organization={The Laboratory for AI-Powered Financial Technologies},
	city={Hong Kong},
	country={China}}
\begin{abstract}
Covariate-dependent uncertainty quantification in simulation-based inference is crucial for high-stakes decision-making but remains challenging due to the limitations of existing methods such as conformal prediction and classical bootstrap, which struggle with covariate-specific conditioning. We propose Efficient Quantile-Regression-Based Generative Metamodeling (E-QRGMM), a novel framework that accelerates the quantile-regression-based generative metamodeling (QRGMM) approach by integrating cubic Hermite interpolation with gradient estimation. Theoretically, we show that E-QRGMM preserves the convergence rate of the original QRGMM while reducing grid complexity from $O(n^{1/2})$ to $O(n^{1/5})$ for the majority of quantile levels, thereby substantially improving computational efficiency. Empirically, E-QRGMM achieves a superior trade-off between distributional accuracy and training speed compared to both QRGMM and other advanced deep generative models on synthetic and practical datasets. Moreover, by enabling bootstrap-based construction of confidence intervals for arbitrary estimands of interest, E-QRGMM provides a practical solution for covariate-dependent uncertainty quantification.
\end{abstract}



\begin{keyword}
conditional uncertainty quantification \sep bootstrap \sep generative metamodeling \sep quantile regression 


\end{keyword}

\end{frontmatter}



\section{Introduction}

Simulation-based decision-making with covariates is important in various domains such as personalized medicine \citep{shen2021ranking}, derivative pricing \citep{jiang2024real}, and risk management \citep{zhang2025conditional}. Recent advances in generative metamodeling \citep{hong2023learning} offer a powerful approach to these problems: by learning the conditional distribution of simulation outputs with respect to the input covariates in the offline stage, given the observed input covariates in the online stage, generative metamodels can produce large-scale conditional random observations on demand, enabling accurate real-time estimation of key distributional estimands (e.g., conditional mean and quantile) to support covariate-informed decisions.

Despite these advances, a significant gap remains: existing generative metamodeling methods do not provide systematic tools for covariate-dependent uncertainty quantification. That is, they do not quantify \emph{how accurate and reliable the generative metamodel’s estimation of the target estimand is}. From a statistical perspective, when a generative metamodel is used for decision-making, it is desirable not only to obtain a point estimate of the estimand, but also to quantify the associated uncertainty through a confidence interval. This is especially important in high-stakes applications, where relying on point estimates without accompanying confidence intervals may expose decision-makers to substantial risks. More importantly, the confidence interval should depend on the covariates in the same way as the estimand estimation does, ensuring coherent covariate-dependent inference.  Standard approaches such as conformal prediction \citep{vovk2005algorithmic,shafer2008tutorial,balasubramanian2014conformal} and the classical bootstrap \citep{stine1985bootstrap,efron1992bootstrap,efron1994introduction} are widely used and have proven effective for uncertainty quantification in many settings, but they are not well suited to our covariate-dependent regime. Conformal prediction provides finite-sample coverage guarantees, yet \citet{lei2014distribution}, \citet{foygel2021limits} and \citet{vovk2012conditional} show that exact conditional coverage is generally unattainable under realistic distributional assumptions. In addition, its use here is limited because the estimand of interest is unobservable, whereas conformal calibration requires access to a suitable notion of ground truth. The classical bootstrap relies on resampling the observed data and typically requires weaker assumptions. However, when covariates are continuous, there may be too few comparable observations at a given covariate value, making conditional bootstrap estimates unstable and, consequently, bootstrap-based confidence intervals infeasible.
 \citet{barton2014quantifying} proposed conditional confidence intervals by retraining ensemble metamodels via bootstrap; however, their approach requires pre-specifying the estimand of interest, which reduces flexibility.

A more effective alternative is to incorporate \emph{generative} metamodels \citep{hong2023learning}, rather than the metamodels tailored to a pre-specified estimand of interest  in \citet{barton2014quantifying}, within a bootstrap framework for conditional uncertainty quantification. By generating large numbers of synthetic conditional observations at any covariate value, generative metamodels enable practitioners to estimate arbitrary estimands of interest on demand during real-time decision-making. This avoids the need to rely on a metamodel tailored to a single pre-specified estimand. Under the bootstrap framework, multiple generative metamodels are trained on bootstrap-resampled datasets, and confidence intervals are constructed from the empirical distribution of the resulting estimates of the estimands across these generative metamodels, improving the flexibility and reliability of conditional inference. A central challenge, however, is computational feasibility: training a generative metamodel can already be time-consuming, and bootstrap inference often requires retraining hundreds of models \citep{wager2014confidence}, which can be prohibitively expensive. Quantile-regression-based generative metamodeling (QRGMM) \citep{hong2023learning} provides accurate conditional distribution estimation but remains computationally costly in this bootstrap setting. This motivates accelerating QRGMM so that bootstrap retraining becomes computationally practical.

In this paper, we propose \emph{Efficient Quantile-Regression-Based Generative Metamodeling} (E-QRGMM), a computationally efficient variant of QRGMM that substantially reduces training cost while retaining accurate approximation of the full conditional quantile process. The key idea is to avoid the QRGMM strategy of fitting quantile regressions on a dense uniform grid. Instead, E-QRGMM fits the quantile regression model on a carefully designed, much sparser set of grid points in the central region, and reconstructs the whole quantile process via cubic Hermite interpolation, rather than the linear interpolation used in QRGMM. By leveraging gradient information, cubic Hermite interpolation supports accurate reconstruction with sparser grids. To make cubic interpolation practical, E-QRGMM further develops an efficient estimator of the quantile-process gradient with respect to the quantile level using a pathwise sensitivity approach \citep{hong2010pathwise}, so that the required gradients can be computed directly from the data.
 E-QRGMM enables fast generation of synthetic conditional observations for any covariate value, which in turn supports flexible estimation of estimands of interest through Monte Carlo evaluation. Crucially, the reduced training cost makes bootstrap-based conditional uncertainty quantification feasible: multiple generative metamodels can be trained on bootstrap-resampled datasets, the estimands of interest can be estimated from each generative metamodel, and confidence intervals can be constructed from the empirical quantiles of these bootstrap estimates. As a result, E-QRGMM provides a computationally viable and broadly applicable solution for covariate-dependent uncertainty quantification in simulation-based decision-making.

 While our work targets simulation-based applications, it connects to the broader field of conditional generative modeling \citep{ajay2022conditional}, which has achieved remarkable success across a wide range of tasks. Notable use cases include image generation \citep{mirza2014conditional}, text generation \citep{brown2020language}, and contextual optimization \citep{liang2024generative}. However, unlike typical machine learning settings, our approach emphasizes (i) distributional fidelity over instance quality, (ii) computational efficiency enabling extensive bootstrap replications, and (iii) low-dimensional outputs, which aligns with common requirements in simulation-based inference.
 
 Throughout the paper, we consider a univariate simulation output \(Y\) and a multivariate covariate vector \(\bx\), and aim to construct valid confidence intervals for scalar estimands of interest \(\psi(Y(\bx))\). Our main contributions are summarized as follows:
\begin{itemize}[noitemsep,topsep=0pt,leftmargin=*]
    \item We propose efficient quantile-regression-based metamodeling (E-QRGMM), which enables the application of cubic Hermite interpolation to quantile process estimation through gradient estimation of quantile functions, thereby achieving significant computational efficiency gains. Consequently, the proposed approach permits efficient implementation of bootstrap-based covariate-dependent confidence interval construction.
    \item Theoretically, we show that E-QRGMM attains the same overall convergence rate as QRGMM while requiring substantially fewer grid points, reducing the number of quantile regressions from \(O(n^{1/2})\) to \(O(n^{1/5})\) for the majority of quantile levels. This delivers a provable computational speedup without sacrificing accuracy in conditional distribution estimation.
    \item Empirically, E-QRGMM substantially outperforms the original QRGMM and other leading deep conditional generative models, offering a better overall trade-off between (i) conditional distribution estimation accuracy, assessed by the Kolmogorov--Smirnov (KS) statistic and the Wasserstein distance (WD), and (ii) training time. Moreover, we demonstrate that the resulting confidence intervals achieve accurate empirical coverage across a range of synthetic and practically motivated datasets and estimands of interest.
\end{itemize}

The rest of our paper is organized as follows. In Section \ref{related work}, we discuss related literature in detail. In Section \ref{problem formulation}, we formally state our goal of constructing confidence intervals conditioned on covariates. In Section \ref{E-QRGMM}, we propose the E-QRGMM framework. In Section \ref{theoretical}, we establish the theoretical foundation for E-QRGMM. In Section \ref{empirical}, numerical experiments empirically demonstrate the efficiency of E-QRGMM and the validity of the bootstrap-based conditional confidence intervals across a variety of scenarios involving both synthetic and real-world datasets. In Section \ref{conclusion}, we conclude our paper and point out a few directions for future research.

\section{Related literature}\label{related work}
Conditional uncertainty quantification is a challenging problem. When covariates are continuously distributed, conditioning on an exact covariate value leads to an event of probability zero, which complicates the direct use of conformal prediction and standard bootstrap methods. This work uses a bootstrap-based uncertainty quantification scheme inspired by \citet{barton2014quantifying} and \citet{10407968}. Our key distinction is the use of generative models, which enables covariate-dependent confidence intervals for a broad class of estimands of interest computed from generated observations, rather than being tailored to a single predefined estimand. A crucial ingredient in this approach is choosing generative models that can approximate the target conditional distribution both efficiently and accurately.

\paragraph{Quantile-regression-based generative metamodeling} QRGMM \citep{hong2023learning} leverages the idea of the inverse transform method to capture the underlying distribution of data. Let $\{(\bx_i,y_i)\}_{i=1}^n$ denote the dataset. Let $F_Y(y \mid \bx)$ be the cumulative distribution function of the univariate variable $Y$ conditioned on the covariate $\bx$ (denoted as $Y(\bx)$). The conditional quantile function is defined by 
\[Q(\tau\mid \bx) \equiv \inf\{y:F_Y(y\mid \bx)\ge \tau\}.\]QRGMM first solves quantile regressions at a set of grid points $\{\tau_j:= j/m \}_{j=1}^{m-1}$, yielding estimates $\hat{Q}(\tau_j\mid\bx)$ for each $\tau_j$. Then for any $\tau \in (0,1)$, the quantile function $\hat{Q}(\tau\mid \bx)$ is obtained by linear interpolation of the values $\{\hat{Q}(\tau_j\mid \bx)\}_{j=1}^{m-1}$. By evaluating \(\hat{Q}(U \mid \bx)\) at random draws from the standard uniform distribution \(U \sim \mathcal{U}(0,1)\), one can generate observations which follow the approximately same distribution as \(Y(\bx)\). Further details can be found in Section~\ref{subsec: qrgmm} and~\ref{appendix:QRGMM}. While QRGMM has demonstrated good accuracy, its training speed leaves substantial room for improvement. In this work, we propose an innovative approach that estimates the gradients $ D(\tau \mid \mathbf{x}) = \frac{d}{d\tau} Q(\tau\mid \mathbf{x}) $ at the quantile levels $\tau$, enabling the use of cubic interpolation in place of linear interpolation for constructing the quantile process $\hat{Q}(\tau\mid\bx)$. This enhancement reduces the number of required grid points and significantly accelerates the training process, while preserving the accuracy of the original method.

\paragraph{Other generative models for approximating conditional distribution} A wide range of conditional generative models can also be employed, as they learn from data to generate the corresponding conditional distribution given the input covariates. Generative adversarial networks (GANs) \citep{goodfellow2014generative}, denoising diffusion implicit models (DDIM) \citep{song2020denoising}, and rectified flow (RectFlow) \citep{liu2022flow} have demonstrated strong capabilities in approximating conditional distributions across tasks such as image generation \citep{mirza2014conditional} and non-convex optimization \citep{liang2024generative}. Although these models are neural-network-based and often entail longer training time, they still offer meaningful benchmarks and references for evaluating conditional generative performance.

In summary, the proposed E-QRGMM addresses the computational challenges associated with training conditional generative models and improves the trade-off between training time and distributional estimation accuracy: under comparable training times, it achieves higher distributional estimation accuracy than QRGMM and many other generative approaches. This enhanced performance enables the effective use of the bootstrap framework for constructing covariate-dependent confidence intervals.

\section{Problem formulation}\label{problem formulation}
This section formalizes the covariate-dependent uncertainty quantification problem and outlines a bootstrap-based procedure for building conditional confidence intervals using generative metamodels.

Let $\bx \in \mathcal{X} \subset \mathbb{R}^p$ denote a $p$-dimensional covariate vector, $Y(\bx)$ be the corresponding univariate simulation output associated with $\bx$. Let $\psi\left(Y(\bx)\right)$ denote a scalar estimand of interest, defined as a functional of the distribution of $Y(\bx)$ and chosen to reflect the underlying decision-making objective.
When no confusion arises, we use the shorthand notation $\psi(\bx)$ to denote $\psi\left(Y(\bx)\right)$. For example, in personalized healthcare, $Y(\bx)$ may represent the predicted life years of a patient with clinical and demographic characteristics $\bx$.
Depending on the clinical context and patient preferences, the estimand of interest $\psi(\bx)$ may take different forms.
In some cases, a patient or decision-maker may focus on the expected outcome, leading to $\psi(\bx)=\mathbb{E}[Y(\bx)]$; in other cases, the primary concern may be the likelihood that the outcome exceeds a clinically meaningful threshold $\tilde{y}$, in which case $\psi(\bx)=\mathbb{P}(Y(\bx)\ge \tilde{y})$. 

Suppose we have a dataset $\{(\bx_i,y_i)\}_{i=1}^n$ where $y_i$ is the realization of $Y(\bx_i)$. Given the real-time observed covariates $\bx=\bx^*$, our goal is to construct a $(1-\alpha)$ confidence interval $[\hat{\psi}_l(\bx^*),\hat{\psi}_u(\bx^*)]$ such that
\[
\mathbb{P}\left( \psi(\bx^*) \in [\hat{\psi}_l(\bx^*), \hat{\psi}_u(\bx^*)] \right) \geq 1 - \alpha,\quad \forall \bx^* \in \mathcal{X}.
\]
To achieve this goal, we generate $B$ bootstrap datasets $\{\mathcal{D}^{(b)}\}_{b=1}^B$ by sampling with replacement from the original dataset, where each $\mathcal{D}^{(b)} = \{(\bx_i^{(b)}, y_i^{(b)})\}_{i=1}^n$. We train a model $\hat{Q}^{(b)}(\tau\mid\bx)$ for every $\mathcal{D}^{(b)}$. Once the covariates $\bx = \bx^*$ are observed, we draw $\{u_k^{(b)}\}_{k=1}^K$ from $\mathcal{U}(0,1)$ and get the output observations $\{\hat{Y}^{(b)}_k(\bx^*):= \hat{Q}^{(b)}(u_k^{(b)}\mid \bx^*)\}_{k=1}^K$. Furthermore, we use these outputs to estimate the target estimands. As an example, if the target estimand is the conditional mean, i.e., $\psi(Y(\bx^*)) = \mathbb{E}[Y(\bx) \mid \bx = \bx^*]$, it can be approximated using the average of the model outputs:
\[
 \hat{\psi}(\hat{Y}^{(b)}(\bx^*)) =   \frac{1}{K} \sum_{k=1}^K \hat{Y}_k^{(b)}( \bx^*).
\]
To simplify notation, we substitute  $\hat{\psi}(\hat{Y}^{(b)}(\bx^*))$ with $\hat{\psi}^{(b)}(\bx^*)$ to denote the estimates of target estimands from the $b$-th bootstrap dataset $\mathcal{D}^{(b)}$ using the aforementioned procedure.

Repeating the procedure across all $B$ bootstrap datasets yields a collection of $B$ independent estimates of $\psi(\bx^*)$, denoted by $\{\hat{\psi}^{(b)}(\bx^*)\}_{b=1}^B$. Let $\hat{\psi}_l(\bx^*)$ and $\hat{\psi}_u(\bx^*)$ be the $\alpha/2$ and $1 - \alpha/2$ empirical quantiles of the set $\{\hat{\psi}^{(b)}(\bx^*)\}_{b=1}^B$, respectively. Then $[\hat{\psi}_l(\bx^*),\hat{\psi}_u(\bx^*)]$ forms the $(1 - \alpha)$ bootstrap percentile confidence interval for $\psi(\bx^*)$.

\section{Efficient quantile-regression-based generative metamodeling}\label{E-QRGMM}
This section presents the proposed E-QRGMM. We first review the original QRGMM construction in Section~\ref{subsec: qrgmm}. We then introduce cubic Hermite interpolation in Section~\ref{subsec:cubic} and develop an efficient gradient estimator needed for the interpolation in Section~\ref{subsection:gradient estimation}. Next, we propose a numerically stable grid design in Section~\ref{subsection:grid point design}. Finally, Section~\ref{subsec: eqrgmm} describes how these components are combined to enable efficient covariate-dependent uncertainty quantification via bootstrap.

To make this bootstrap-based pipeline practically feasible, the underlying conditional distribution model must be both accurate and computationally efficient, since it will be retrained repeatedly across bootstrap resamples. Accordingly, we adopt a simple yet effective specification based on linear quantile regression, $\hat{Q}(\tau\mid \bx)=\bx^\top\hat{\bbeta}(\tau)$. Throughout the paper, we assume the following linear model holds, which simplifies the gradient estimation in Section~\ref{subsection:gradient estimation} and the theoretical analysis in Section~\ref{theoretical}.

\begin{assumption}\label{assumption:linear}
	Suppose that $\bbeta(\tau) \in \mathbb{R}^p$ and the true conditional quantile function takes the form
	$     Q(\tau \mid \bx) = \bx^\top \bbeta(\tau) , \quad \forall \tau \in (0, 1).
	$
	In addition, we assume that $\bbeta(\tau)$ is continuous in $\tau$.
\end{assumption}
This assumption can be generalized to nonlinear forms by replacing \(\mathbf{x}\) with complex basis functions \(\mathbf{b}(\mathbf{x})\), i.e., \(Q(\tau \mid \mathbf{x}) = \mathbf{b}(\mathbf{x})^\top \boldsymbol{\beta}(\tau)\), and our theoretical results continue to hold under this extension. In practice, even when the model is misspecified, our experiments in Section~\ref{subsec: practical data} show that our method still performs well empirically for practical data.
 
Given a dataset $\{(\bx_i,y_i)\}_{i=1}^{n}$, the quantile regression parameter vector $\bbeta(\tau)$ can be obtained by solving the following optimization problem \citep{koenker1978regression}:
\begin{equation}\label{pinball loss}
    \hat{\bbeta}(\tau) \equiv \mathop{\arg\min}\limits_{\beta \in \mathbb{R}^p} \, \sum_{i=1}^n \rho_\tau(y_i -  \bx_i^\top \bbeta).
\end{equation}
where the pinball loss function is defined as \(
\rho_\tau(u)=\big(\tau-\mathbbm{1}\{u\le 0\}\big)u,
\) with $\mathbbm{1}\{\cdot\}$ denoting the indicator function. Intuitively, $\rho_\tau$ penalizes positive and negative residuals asymmetrically. This asymmetry encourages the fitted value to sit at a level where roughly a $\tau$ fraction of outcomes fall below it and a $(1-\tau)$ fraction fall above it. As a result, minimizing this loss targets the conditional $\tau$-quantile. Moreover, problem \eqref{pinball loss} can be formulated as a linear program and thus solved efficiently using standard optimization solvers \citep{koenker2005}.

\subsection{Quantile-regression-based generative metamodeling}\label{subsec: qrgmm}
Our work relates closely to \citet{hong2023learning}, the problem defined by Eq.~\eqref{pinball loss} is solved at a set of grid points \(\{\tau_j=j/m\}_{j=1}^{m-1} \), obtaining the corresponding estimates \( \{ \hat{Q}(\tau_j\mid \bx) = \bx^\top\hat{\bbeta}(\tau_j)\}_{j=1}^{m-1}\). The estimator \(\hat{Q}(\tau\mid \bx)\) for any \(\tau \in (0,1)\) is then obtained via linear interpolation over \( \{ \hat{Q}(\tau_j\mid \bx) \}_{j=1}^{m-1}\), yielding the original QRGMM quantile process estimator \(\hat{Q}_O(\tau\mid\bx)\) defined as
\begin{equation}\label{eq:qrgmm_linear_interp}
\hat{Q}_O(\tau \mid \bx) =
\begin{cases}
\hat{Q}(\tau_1 \mid \bx), & \tau < \tau_1, \\[2pt]
\hat{Q}(\tau_j \mid \bx) + m(\tau-\tau_j)\bigl[\hat{Q}(\tau_{j+1}\mid \bx)-\hat{Q}(\tau_j\mid \bx)\bigr],
& \tau \in [\tau_j,\tau_{j+1}), \\[2pt]
\hat{Q}(\tau_{m-1} \mid \bx), & \tau \ge \tau_{m-1}.
\end{cases}
\end{equation}
Given a covariate value \(\bx=\bx^*\), QRGMM generates observations by drawing \(u\sim\mathcal{U}(0,1)\) and returning \(\hat{Q}_O(u\mid\bx^*)\); see~\ref{appendix:QRGMM} for further details.

\subsection{Cubic Hermite interpolation}\label{subsec:cubic}
In contrast to estimating the conditional quantile function on a uniform grid and applying linear interpolation, we adopt cubic Hermite interpolation over a more sparse grid, substantially reducing the computational burden. For $\tau \in (\tau_j,\tau_{j+1})$, assume we have the gradient estimates $\hat{D}(\tau_j\mid\bx)$ and $\hat{D}(\tau_{j+1}\mid\bx)$. The cubic Hermite interpolator is defined as

\begin{equation}\label{eq:cubic_hermite}
\begin{aligned}
\hat{Q}_C(\tau \mid \bx) ={}& h_{00}(\xi)\hat{Q}_j
+ h_{10}(\xi)(\tau_{j+1}-\tau_j)\hat{D}_j \\
&+ h_{01}(\xi)\hat{Q}_{j+1}
+ h_{11}(\xi)(\tau_{j+1}-\tau_j)\hat{D}_{j+1}.
\end{aligned}
\end{equation}
where \( \xi = \dfrac{\tau - \tau_j}{\tau_{j+1} - \tau_j} \), \( \hat{Q}_j = \hat{Q}(\tau_j \mid \bx) \), \( \hat{D}_j = \hat{D}(\tau_j \mid \bx) \), and likewise for \( \tau_{j+1} \).

The Hermite basis functions are
\[
\begin{aligned}
h_{00}(\xi) &= 2\xi^3 - 3\xi^2 + 1, &
h_{01}(\xi) &= -2\xi^3 + 3\xi^2, \\
h_{10}(\xi) &= \xi^3 - 2\xi^2 + \xi, &
h_{11}(\xi) &= \xi^3 - \xi^2.
\end{aligned}
\]

\subsection{Gradient estimation}\label{subsection:gradient estimation} 

To implement the cubic Hermite interpolation defined in Eq.~\eqref{eq:cubic_hermite}, it is essential to estimate the gradient of the conditional quantile function with respect to the quantile level $\tau$. However, gradient estimation for the quantile regression process is nontrivial due to the presence of non-differentiable indicator functions in the pinball loss used in quantile regression. 
To address this issue, we adopt the pathwise sensitivity estimation approach proposed by \citet{hong2010pathwise} to overcome this obstruction. Building on this approach, we derive an efficient estimator for the gradient $D(\tau \mid \bx) = \frac{d}{d\tau} Q(\tau \mid \bx)$, as summarized in Proposition~\ref{pro:gradient_estimation}. A detailed derivation is provided in  \ref{appendix:gradient_proof}.

\begin{proposition}[Gradient estimation]\label{pro:gradient_estimation}
Under Assumption~\ref{assumption:linear}, and Assumptions \ref{ass:pathwise} and \ref{ass:residual_smoothness} in~\ref{app:assumptions}, the gradient $D(\tau\mid \bx)$ satisfies
\[
D(\tau \mid \mathbf{x}) = \frac{d}{d\tau} Q(\tau \mid \mathbf{x}) = \mathbf{x}^{\top} \frac{d}{d\tau} \bbeta(\tau) = \mathbf{x}^{\top} \Lambda^{-1}(\tau) \mathbb{E}[\bx], 
\]
where \(\Lambda(\tau) = \lim_{\delta \to 0^+} \frac{1}{2\delta} \mathbb{E}\left[ \bx\bx^\top \cdot \mathbbm{1}\{-\delta < Y - \bx^\top \bbeta(\tau) < \delta\} \right].
\)
\end{proposition}
Given a covariate value $\bx = \bx^*$, $D(\tau\mid\bx^*)$ can be directly estimated from data $\{(\bx_i,y_i)\}_{i=1}^n$ by
\begin{equation}\label{eq:gradient estimation}
    \hat{D}(\tau \mid \mathbf{x}^*) = \bx^{*\top}\hat{\Lambda}^{-1}(\tau) \bar{\bx},
\end{equation}
with \( \bar{\bx} = \frac{1}{n} \sum_{i=1}^n \bx_i, \hat{\Lambda}(\tau) = \frac{1}{2 \delta_n} \frac{1}{n} \sum_{i=1}^n \bx_i \bx_i^\top \cdot \mathbbm{1}\{-\delta_n < y_i - \bx_i^\top \hat{\bbeta}(\tau) < \delta_n\},\)
where $\hat{\bbeta}(\tau)$ is the empirical estimator of the $\tau$-quantile regression coefficients, obtained by minimizing the pinball loss in Eq.~\eqref{pinball loss}.
For gradient estimation, we set the smoothing parameter $\delta_n = O(n^{-1/5})$, following \citet{hong2010pathwise}.

\subsection{Grid point design}\label{subsection:grid point design}

Although we propose an effective method for gradient estimation in Section~\ref{subsection:gradient estimation}, the limited data in the tails may lead to highly unstable gradient estimates at extreme quantile levels $\tau $ approaching 0 or 1, see  \ref{app:tail gradient} for details. To address the numerical instability, we propose a new grid point design \(\mathcal{T}(m)\) parameterized by \(m\). It employs a sparse grid with spacing significantly larger than \(1/m\) only within the specified central region \([\tau_{\mathsf{L}}, \tau_{\mathsf{U}}]\), where \(\tau_{\mathsf{L}}\) and \(\tau_{\mathsf{U}}\) are fixed constants, while retaining the original uniform \(1/m\)-spaced grid in the tails. Specifically, we define a fine grid in the tail and a coarse grid in the main region as follows:
\[
\begin{aligned}
\mathcal{T}_O(m) &= \left\{ {\tau}_j = \frac{j}{m} : \frac{j}{m} \in (0,1) \setminus [\tau_{\mathsf{L}}, \tau_{\mathsf{U}}], j = 1,2,\ldots,m-1 \right\}; \\
\mathcal{T}_C(m) &= \left\{ \tau_k = \tau_{\mathsf{L}} + \frac{k}{m'} (\tau_{\mathsf{U}} - \tau_{\mathsf{L}}) : k = 0, 1, \dots, m' \right\}, \quad m' = \lceil c \cdot m^{2/5}\rceil.
\end{aligned}
\]
where $\lceil \cdot \rceil$ means rounding up, and $c$ is a constant to avoid near-empty sets when $m$ is small, typically set to $2$. The number of grid points in the main region, $m' = O(m^{2/5})$, is significantly reduced compared to the original setting with order $O(m)$, its choice is supported by the theoretical analysis in Section~\ref{theoretical}. The complete set of grid points is then given by the union $\mathcal{T}_{\text{raw}} = \mathcal{T}_O \cup \mathcal{T}_C$ and we define the final grid \(\mathcal{T}(m) = \{ \tau_1 < \tau_2 < \dots < \tau_J \}\) as the increasing rearrangement of \(\mathcal{T}_{\text{raw}}\). And we use $\{\hat{Q}(\tau_j \mid \bx)\}_{j=1}^J$ and $\{\hat{D}(\tau_j \mid \bx):j\in\{1,\ldots,J\},\tau_j\in[\tau_\mathsf{L},\tau_\mathsf{U}]\}$ to denote the estimated conditional quantiles and the needed gradients, respectively.

\begin{algorithm}[!t]
\caption{Bootstrap Confidence Interval for General Estimands of Interest}
\label{alg:bootstrap_general}
\begin{algorithmic}[1]
\State \textbf{Input:} Dataset $\mathcal{D} = \{(\bx_i, y_i)\}_{i=1}^n$; grid point design $\mathcal{T}(m)$; auxiliary parameters for grid point design $c,\tau_\mathsf{L},\tau_\mathsf{U}$; smoothing parameter $\delta_n$; number of bootstraps $B$; number of generated observations per bootstrap $K$;  evaluation covariate point $\bx^*$; significance level $\alpha$; estimator $\hat{\psi}(\cdot)$.

\For{$b = 1$ to $B$}
    \State Sample with replacement from $\mathcal{D}$ to get  $\mathcal{D}^{(b)}:=\{(\bx_i^{(b)},y_i^{(b)})\}_{i=1}^n$ ,
\State Solve Eq.~\eqref{pinball loss} to get $\{\hat{\bbeta}(\tau_j\mid\bx)\}_{j=1}^{J}$ for $\tau_j \in \mathcal{T}(m)$ on $\mathcal{D}^{(b)}$,
\State Calculate $\{\hat{D}(\tau_j \mid \bx):j\in\{1,\ldots,J\},\tau_j\in[\tau_\mathsf{L},\tau_\mathsf{U}]\}$  by Eq.~\eqref{eq:gradient estimation} on $\mathcal{D}^{(b)}$,
    \State Generate $\hat{Y}_k^{(b)}(\bx^*) := \hat{Q}^{(b)}(u_k^{(b)} \mid \bx^*)$ by Eq.~\eqref{eq:mixed_interp} with $u_k^{(b)} \stackrel{\text{i.i.d.}}{\sim} \mathcal{U}(0,1)$, $k = 1, \dots, K$,
    \State Compute estimates: $\hat{\psi}^{(b)}(\bx^*) := \hat{\psi}\left(\{ \hat{Y}_k^{(b)}(\bx^*) \}_{k=1}^{K} \right)$.
\EndFor

\State Compute confidence bounds:
\[
\hat{\psi}_l(\bx^*) := q_{\alpha/2} \left( \left\{ \hat{\psi}^{(b)}(\bx^*) \right\}_{b=1}^{B} \right), \quad
\hat{\psi}_u(\bx^*) := q_{1 - \alpha/2} \left( \left\{ \hat{\psi}^{(b)}(\bx^*) \right\}_{b=1}^{B} \right).
\]
\State \textbf{Output:} The $(1-\alpha)$ confidence interval $[\hat{\psi}_l(\bx^*), \hat{\psi}_u(\bx^*)]$.
\end{algorithmic}
\end{algorithm}

\subsection{Efficient covariate-dependent uncertainty quantification}\label{subsec: eqrgmm}

Denote $\hat{Q}_O(\tau \mid \bx)$ in Eq.~\eqref{eq:qrgmm_linear_interp} as the original QRGMM estimator. By employing $\hat{Q}_O(\tau\mid\bx)$ for tail quantile levels $\tau \notin [\tau_{\mathsf{L}}, \tau_{\mathsf{U}}]$ and $\hat{Q}_C(\tau\mid\bx)$ defined by Eq.~\eqref{eq:cubic_hermite} over the central quantile levels $\tau \in[\tau_{\mathsf{L}}, \tau_{\mathsf{U}}]$, E-QRGMM is able to avoid numerical instability while still achieving substantial computational savings. 
Formally, for $ \tau\in(0,1)$, the E-QRGMM is defined as:
\begin{equation}
\hat{Q}(\tau \mid \bx) := 
\begin{cases}
\hat{Q}_C(\tau\mid \bx), & \text{if } \tau \in [\tau_{\mathsf{L}}, \tau_{\mathsf{U}}], \\[6pt]
\hat{Q}_O(\tau \mid \bx), & \text{if } \tau \not\in [\tau_{\mathsf{L}}, \tau_{\mathsf{U}}].
\end{cases}
\label{eq:mixed_interp}
\end{equation}

The design in Eq.~\eqref{eq:mixed_interp} leverages the strengths of both methods: $\hat{Q}_C$ delivers efficiency in the main region $[\tau_{\mathsf{L}}, \tau_{\mathsf{U}}]$, while $\hat{Q}_O$ ensures robustness and consistency in the tails. This combination allows the E-QRGMM $\hat{Q}(\tau \mid \bx)$ to achieve better overall performance.

E-QRGMM supports conditional observation generation, facilitating Monte Carlo estimation of a broad class of covariate-dependent estimands at a given covariate value \(\bx^*\). Given the observations \(\{\hat{Y}_k(\bx^*)\}_{k=1}^K\) generated by E-QRGMM, we denote the estimator for an estimand of interest by \(\hat{\psi}(\cdot)\). A few illustrative examples are as follows:
\begin{itemize}[noitemsep,topsep=0pt,leftmargin=*]
    \item \textbf{Conditional mean} $\mathbb{E}[Y \mid \bx^*]$: \quad $\hat{\psi}(\{\hat{Y}_k(\bx^*)\}) = \frac{1}{K} \sum_{k=1}^{K} \hat{Y}_k(\bx^*)$;
    \item \textbf{Conditional quantile} $q_\tau(Y \mid \bx^*)$: \quad $\hat{\psi}(\{\hat{Y}_k(\bx^*)\})= q_{\tau}(\{\hat{Y}_k(\bx^*)\}_{k=1}^K)$;
    \item \textbf{Survival function} $\mathbb{P}(Y>\tilde{y} \mid \bx^*)$: \quad $\hat{\psi}(\{\hat{Y}_k(\bx^*)\})= \frac{1}{K} \sum_{k=1}^{K} \mathbbm{1}\{ \hat{Y}_k(\bx^*) > \tilde{y} \}$.
\end{itemize}

More importantly, the computational efficiency of E-QRGMM makes repeated refitting feasible, rendering it well suited for bootstrap retraining and, consequently, covariate-dependent uncertainty quantification. In particular, recomputing the estimands of interest across bootstrap replicates yields an empirical sampling distribution, from which \((1-\alpha)\) confidence intervals can be constructed. Algorithm~\ref{alg:bootstrap_general} summarizes this bootstrap-based procedure.

\section{Theoretical analysis}\label{theoretical}

This section establishes the theoretical guarantees for E-QRGMM. Our objective is to construct a cubic Hermite interpolation function, denoted as \( \hat{Q}_C(\tau \mid \mathbf{x}) \), over the interval \( [\tau_{\mathsf{L}}, \tau_{\mathsf{U}}] \), where \( 0 < \tau_{\mathsf{L}} < \tau_{\mathsf{U}} < 1 \). This interpolation is based on estimated quantile values \( \hat{Q}(\tau_j \mid \mathbf{x}) \) and their derivative estimates \( \hat{D}(\tau_j \mid \mathbf{x}) \) at grid points \( \tau_j \in [\tau_{\mathsf{L}}, \tau_{\mathsf{U}}] \). 
Additionally, let \( Q_C(\tau \mid \mathbf{x}) \) represent the cubic Hermite interpolation function constructed using the true quantile values \( Q(\tau_j\mid \mathbf{x})) \) and gradients \( D(\tau_j\mid \mathbf{x})) \) at grid points \( \tau_j \in [\tau_{\mathsf{L}}, \tau_{\mathsf{U}}] \). Our objective is to rigorously analyze the error \( \hat{Q}_C(\tau \mid \mathbf{x}) - Q(\tau \mid \mathbf{x}) \) within \( [\tau_{\mathsf{L}}, \tau_{\mathsf{U}}] \). We decompose it into the interpolation error \( Q_C(\tau \mid \mathbf{x}) - Q(\tau \mid \mathbf{x}) \) (arising from the cubic Hermite approximation between grid points) and the estimation error \( \hat{Q}_C(\tau \mid \mathbf{x}) - Q_C(\tau \mid \mathbf{x}) \) (stemming from the differences between \( \hat{Q}(\tau_j \mid \mathbf{x}) \) and \( Q(\tau_j \mid \mathbf{x}) \), as well as \( \hat{D}(\tau_j \mid \mathbf{x}) \) and \( D(\tau_j \mid \mathbf{x}) \)), and establish the theory of its asymptotic behavior. Specifically, we bound the interpolation error, the quantile regression estimation error, and the gradient estimation error in Lemmas~\ref{lemma:interpolation_error}--\ref{lemma:gradient_estimation_error}, respectively, and finally combine these results to establish the overall convergence rate in Theorem~\ref{thm: rate}.

To facilitate direct comparison with the original QRGMM results and maintain consistency with the QRGMM framework, our analysis begins with the same grid setting as QRGMM: the interval \( [0,1] \) is uniformly partitioned into \( m \) equal subintervals, with grid points defined as \( \tau_j = j/m \) for \( j = 1, \dotsc, m-1 \). Under this configuration, we study the asymptotic behavior of the total error within \( [\tau_{\mathsf{L}}, \tau_{\mathsf{U}}] \). This approach provides practical guidance on optimally selecting the number of grid points \( m \) relative to the number of data points \( n \) within \( [\tau_{\mathsf{L}}, \tau_{\mathsf{U}}] \) to achieve efficient and accurate cubic interpolation in applications. 

Before stating the main theorem, we establish three lemmas that underpin the error analysis: one for the cubic Hermite interpolation error, one for the quantile regression estimation error, and one for the gradient estimation error. The required technical assumptions are standard in the related literature. Specifically, Assumption~\ref{ass:smoothness} characterizes the smoothness of the quantile function, Assumption~\ref{ass:quantile_regression} imposes regularity conditions for quantile regression, and Assumptions~\ref{ass:pathwise}--\ref{ass:moment_diff} facilitate gradient estimation. For completeness and to improve readability, we defer the detailed assumptions to~\ref{app:assumptions}, and present the proofs of Lemmas~\ref{lemma:interpolation_error}, \ref{lemma:gradient_estimation_error} and Theorem~\ref{thm: rate} in~\ref{app:E-QRGMM proofs}.

\begin{lemma}[Cubic Hermite Interpolation Error]\label{lemma:interpolation_error}
	Under Assumption~\ref{assumption:linear}, and Assumption~\ref{ass:smoothness} in~\ref{app:assumptions}, for any fixed \( \mathbf{x}^* \), the true cubic Hermite interpolant \( Q_C(\tau \mid \mathbf{x}^*) \) satisfies 
	\[
	\sup_{\tau \in [\tau_{\mathsf{L}}+\frac{1}{m} , \tau_{\mathsf{U}}-\frac{1}{m}]} \left| Q_C(\tau \mid \mathbf{x}^*) - Q(\tau \mid \mathbf{x}^*) \right| \leq \frac{M}{384}\cdot \frac{1}{m^4}  = O\left( \frac{1}{m^4} \right),
	\]
	where $M= \sup_{\xi \in [\tau_{\mathsf{L}} , \tau_{\mathsf{U}}]} \left| Q^{(4)}(\xi \mid \mathbf{x}^*) \right|$.
\end{lemma}

The focus on the interval \([\tau_{\mathsf{L}}, \tau_{\mathsf{U}}]\) for cubic Hermite interpolation is also motivated by the need to ensure that \(M\) remains bounded. Without restricting \(\tau\) to this interval, as the grid size \(m \to \infty\), the endpoints of the grid would approach 0 or 1. For nonbounded distributions, this could lead to the fourth derivative \(Q^{(4)}(\tau \mid \mathbf{x})\) becoming unbounded, causing \(M\) to diverge to infinity. To maintain the applicability of our method, we therefore confine the use of cubic Hermite interpolation to the interval \([\tau_{\mathsf{L}}, \tau_{\mathsf{U}}]\).

\begin{lemma}[Quantile Regression Estimation Error]\label{lemma:qr_estimation_error}
	Under Assumption~\ref{assumption:linear}, and Assumptions~\ref{ass:smoothness} and \ref{ass:quantile_regression} in~\ref{app:assumptions}, for any fixed \( \mathbf{x}^* \), as $n\to\infty$, the quantile estimator \( \hat{Q}(\tau_j \mid \mathbf{x}^*) \) at grid points \( \tau_j \in [\tau_{\mathsf{L}}, \tau_{\mathsf{U}}] \) satisfies
	\[
	\max_{\tau_j \in [\tau_{\mathsf{L}}, \tau_{\mathsf{U}}]} \left| \hat{Q}(\tau_j \mid \mathbf{x}^*) - Q(\tau_j \mid \mathbf{x}^*) \right| = O_p\left( \frac{1}{\sqrt{n}} \right).
	\]
\end{lemma}

Here, $O_p(\cdot)$ denotes boundedness in probability with respect to the sampling variability of the data. Specifically, $\hat{Q}(\tau_j\mid \bx^*)$ is a random estimator dependent on the sample.
Analogously, the stochasticity in Lemma~\ref{lemma:gradient_estimation_error} also stems from the data randomness.
The result of Lemma~\ref{lemma:qr_estimation_error}
follows directly from Proposition~3 of \citet{hong2023learning}, which establishes a uniform $\sqrt{n}$-rate of convergence for quantile regression estimators at grid points over
$[\tau_{\mathsf{L}}, \tau_{\mathsf{U}}]$.

\begin{lemma}[Gradient Estimation Error]\label{lemma:gradient_estimation_error}
	Under Assumption~\ref{assumption:linear}, and Assumptions~\ref{ass:pathwise}--\ref{ass:moment_diff} in~\ref{app:assumptions}, for any fixed \( \mathbf{x}^* \), as $n\to\infty$, the gradient estimator \( \hat{D}(\tau_j \mid \mathbf{x}^*) \) at grid points \( \tau_j \in [\tau_{\mathsf{L}}, \tau_{\mathsf{U}}] \) satisfies  
	\[
	\max_{\tau_j \in [\tau_{\mathsf{L}}, \tau_{\mathsf{U}}]} \left| \hat{D}(\tau_j \mid \mathbf{x}^*) - D(\tau_j \mid \mathbf{x}^*) \right| = O_p\left( n^{-3/10} \right),
	\]
\end{lemma}

With the above Lemmas~\ref{lemma:interpolation_error}-\ref{lemma:gradient_estimation_error}, we are now ready to present the convergence rate of the \mbox{E-QRGMM estimator}. The proof of the theorem is deferred to the  \ref{app:E-QRGMM proofs}.

\begin{theorem}[Convergence Rate of E-QRGMM]\label{thm: rate}
	Under Assumption~\ref{assumption:linear}, and Assumptions~\ref{ass:smoothness}--\ref{ass:moment_diff} in~\ref{app:assumptions}, for any fixed covariate \( \mathbf{x}^* \), as $n, m\to\infty$, we have 
\[
\sup_{\tau \in [\tau_{\mathsf{L}}+\frac{1}{m},\, \tau_{\mathsf{U}}-\frac{1}{m}]}
\left| \hat{Q}_C(\tau \mid \mathbf{x}^*) - Q(\tau \mid \mathbf{x}^*) \right|
= O_p\!\left( \frac{1}{\sqrt{n}} \right)
+ O_p\!\left( \frac{n^{-3/10}}{m} \right)
+ O\!\left( \frac{1}{m^4} \right).
\]

\end{theorem}
The use of cubic Hermite interpolation improves the convergence rate of the interpolation error within the central quantile region $[\tau_{\mathsf{L}}, \tau_{\mathsf{U}}]$, while preserving the global convergence rate outside this region as in the original QRGMM framework. Consequently, all other theoretical properties of QRGMM remain intact. The improvement is formalized in Theorem~\ref{thm: rate}, which demonstrates that to attain the optimal rate of total error under balanced scaling within $[\tau_{\mathsf{L}}, \tau_{\mathsf{U}}]$, it suffices to choose the number of quantile grid points as $m = O(n^{1/5})$, as opposed to the original requirement $m = O(n^{1/2})$ in QRGMM. Under this choice, the three components of the total error become:
\[
O_{\mathbb{P}}\left( \frac{1}{\sqrt{n}} \right) \text{ (dominant term)}, \quad O_{\mathbb{P}}\left( \frac{n^{-3/10}}{n^{1/5}} \right) = O_{\mathbb{P}}\left( \frac{1}{\sqrt{n}} \right), \quad O\left( \frac{1}{n^{4/5}} \right).
\]
The overall error remains of order $O_{\mathbb{P}}(n^{-1/2})$, as in QRGMM, while this improvement yields a substantial reduction in computational complexity. In the original QRGMM, the number of grid points is $m$, while in the E-QRGMM, the number of grid points in $\mathcal{T}(m)$ is approximately
\[
m \cdot 2\tau_{\mathsf{L}} + c \cdot m^{2/5} \cdot (1 - 2\tau_{\mathsf{L}}),
\]
for some constant $c > 0$ and a symmetric setting $\tau_{\mathsf{L}}=1-\tau_{\mathsf{U}}$, reflecting the reduced cost of implementing quantile regressions in the central region.  For example, if $\tau_{\mathsf{L}} = 0.05$, then as $n$ and $m$ increase, the number of quantile regressions required by E-QRGMM converges to approximately 10\% of that required by the original QRGMM, yielding a significant gain in computational efficiency without compromising statistical accuracy.

\section{Numerical experiments}\label{empirical}

This section reports numerical experiments evaluating E-QRGMM. Section~\ref{subsec: exp setup} describes the experimental setup. Section~\ref{subsec: synthetic exp} presents  results on synthetic datasets, including comparisons with QRGMM and deep conditional generative baselines, as well as validation of bootstrap-based confidence intervals. Section~\ref{subsec: practical data} studies a practically motivated inventory management dataset. All experiments presented in Section \ref{empirical} were performed on a system equipped with two Intel Xeon Gold 6248R CPUs (each featuring 24 cores) and 256GB of RAM. The source codes for reproducing all experiments are publicly available \footnote{\url{https://github.com/Name-less-King/E-QRGMM-Efficient-Generative-Metamodeling-for-Covariate-Dependent-Uncertainty-Quantification}}.

\subsection{Experimental Setup}\label{subsec: exp setup}
\paragraph{Estimands of interest} We consider three commonly used and practically relevant estimands: the conditional mean \(\mathbb{E}[Y\mid \bx]\), the conditional \(80\%\) quantile \(q_{0.8}(Y\mid \bx)\), and the survival function \(P(Y>\tilde{y}\mid \bx)\). These estimands capture central tendency and upper-tail behavior of the conditional distribution. In our experiments, the threshold \(\tilde{y}\) is selected as the conditional \(80\%\) quantile.
\paragraph{Evaluation metrics}
We measure computational efficiency by training time. The quality of conditional distribution approximation is assessed by the Kolmogorov--Smirnov (KS) statistic \citep{massey1951kolmogorov} and the Wasserstein distance (WD) \citep{villani2021topics}. The reliability of the bootstrap confidence intervals is evaluated by empirical coverage and average width.

For the generated conditional CDF $F_{\hat{Y}}(\cdot\mid\bx)$ and the target conditional CDF $F_Y(\cdot\mid\bx)$, KS and WD are defined as follows.
\begin{align*}
\text{KS} &:= \sup_{y\in\mathbb{R}} \bigl|F_{\hat{Y}}(y\mid\bx)-F_Y(y\mid\bx)\bigr|, \\
\text{WD} &:= \int_{\mathbb{R}} \bigl|F_{\hat{Y}}(y\mid\bx)-F_Y(y\mid\bx)\bigr|\,dy .
\end{align*}
KS and WD quantify discrepancies between the generated and target conditional distributions in complementary ways. The KS statistic measures the maximum vertical gap between conditional CDFs and is particularly sensitive to localized deviations, while WD aggregates the discrepancy over the entire support and captures global distributional mismatch. Since the estimands of interest in this work are functionals of the conditional distribution, improved distributional fit in KS/WD typically translates into more accurate estimation of these estimands.

To validate Algorithm~\ref{alg:bootstrap_general}, we run $N$ independent replications and obtain intervals
$\{[\hat{\psi}_l^i(\bx),\,\hat{\psi}_u^i(\bx)]\}_{i=1}^N$. We report
\begin{align*}
\text{Coverage} &:= \frac{1}{N}\sum_{i=1}^N \mathbbm{1}\!\left\{\psi(Y(\bx))\in\bigl[\hat{\psi}_l^i(\bx),\,\hat{\psi}_u^i(\bx)\bigr]\right\}, \\
\text{Width} &:= \frac{1}{N}\sum_{i=1}^N \bigl(\hat{\psi}_u^i(\bx)-\hat{\psi}_l^i(\bx)\bigr).
\end{align*}
Coverage measures calibration by quantifying how often the constructed intervals contain the true estimand, while Width summarizes interval informativeness, with narrower intervals being preferable when coverage is maintained.

\paragraph{Baseline Methods}
We compare against quantile-regression-based generative metamodeling (QRGMM) \citep{hong2023learning}, an accurate baseline for conditional distribution learning which is closely related to our approach. We also consider representative deep conditional generative models, including generative adversarial nets (GAN) \citep{goodfellow2014generative, mirza2014conditional}, denoising diffusion implicit models (DDIM) \citep{song2020denoising}, and rectified flow (RectFlow) \citep{liu2022flow}. These methods are highly expressive and have demonstrated strong performance in modeling complex conditional distributions across a broad range of tasks, and thus provide meaningful and competitive benchmarks in our experiments.

\subsection{Synthetic experiments}\label{subsec: synthetic exp}

We assume a known marginal distribution for the covariates $\bx$. Let $\bx=(1,x_1,x_2,x_3)^\top\in\mathbb{R}^4$ with independent components $x_1\sim\mathcal{U}(0,10)$, $x_2\sim\mathcal{U}(-5,5)$, and $x_3\sim\mathcal{U}(0,5)$. This design yields heterogeneous conditional distributions across covariate values while maintaining a controlled data-generating mechanism. We consider three conditional distributions for $Y\mid \bx$:
\[
\begin{array}{l}
\textbf{Normal: } Y \mid \bx \sim \mathcal{N}(\mu(\bx), \sigma(\bx)); \\
\textbf{Halfnormal: } Y \mid \bx \sim \mathcal{HN}(\mu(\bx), \sigma(\bx)); \\
\textbf{Student's t: } Y \mid \bx \sim t_\nu(\mu(\bx), \sigma(\bx)).
\end{array}
\]
where the location parameter $\mu(\bx)=5+x_1+2x_2+0.5x_3$, the scale parameter $\sigma(\bx)=1+0.1x_1+0.2x_2+0.05x_3$, and the degrees of freedom for the $t$ distribution is $\nu=5$. In these settings, the linear model Assumption~\ref{assumption:linear} holds.
These choices are designed to test performance under different tail behaviors and symmetry properties. For all the experiments in Section \ref{subsec: synthetic exp}, we use a training set $\{(\bx_i,y_i)\}_{i=1}^n$ with $n=10^4$ to fit generative models, and generate $K=10^5$ observations with covariate $ \bx^* = (1,4,-1,3)$ from every model to evaluate performance. All results are presented as mean values with standard errors derived from $N=100$ independent experimental replications. Implementation details for all models are provided in~\ref{app:detailed exp}.

\subsubsection{E-QRGMM Improves Computational Efficiency over QRGMM}

\begin{figure}[!t]
    \centering
    \includegraphics[width=0.9\linewidth]{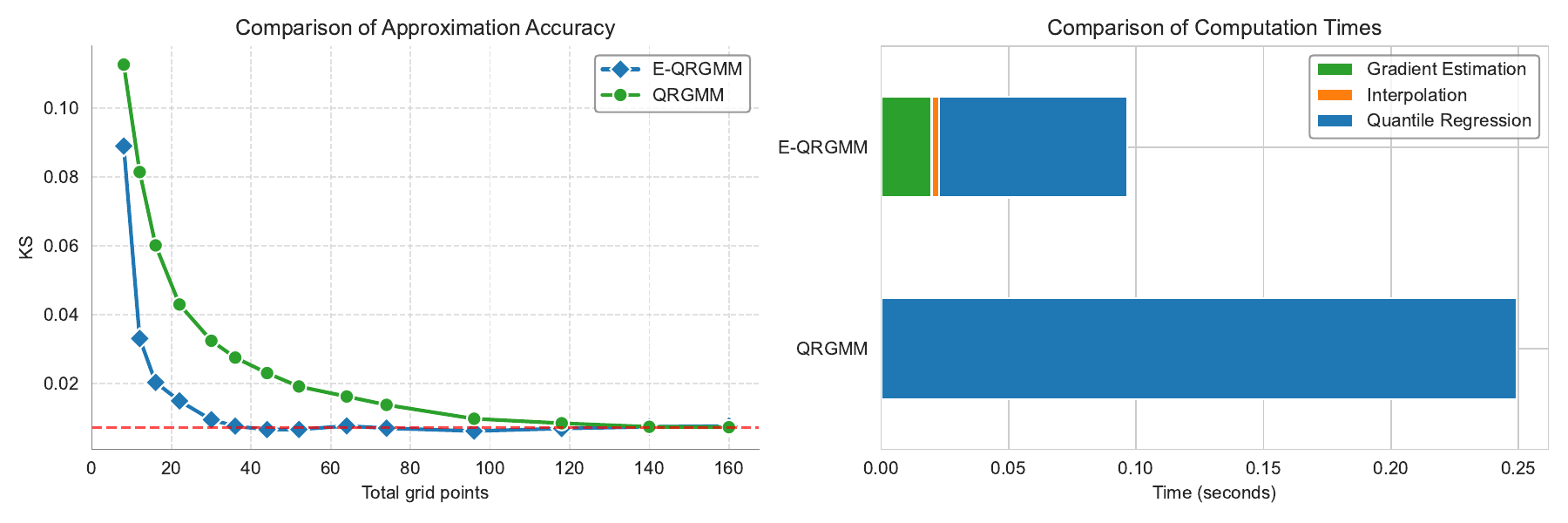}
    \caption{Comparison of E-QRGMM and QRGMM on normal test distribution with $n=10^4$. \textbf{Left:} E-QRGMM achieves the same distributional accuracy with fewer grid points. \textbf{Right:} In E-QRGMM, quantile regression remains the dominant cost, and total time is much lower than in QRGMM.}
    \label{fig:E-QRGMM and QRGMM}
\end{figure}

We first conduct ablation studies to compare our proposed E-QRGMM method with QRGMM under the same total number of grid points. As shown in Figure~\ref{fig:E-QRGMM and QRGMM}, for comparable distribution estimation accuracy (quantified by the KS statistic), E-QRGMM requires substantially fewer grid points than QRGMM. Equivalently, for a fixed grid budget, E-QRGMM attains better accuracy due to the more informative interpolation scheme in the central region. Moreover, the additional computation introduced by gradient estimation and cubic Hermite interpolation accounts for only a small portion of the total runtime in practice, since these operations are carried out only on the sparse central grid \(\mathcal{T}_C\) and do not scale with the full uniform grid size.

\begin{figure}[!t]
    \centering
    \includegraphics[width=0.9\linewidth]{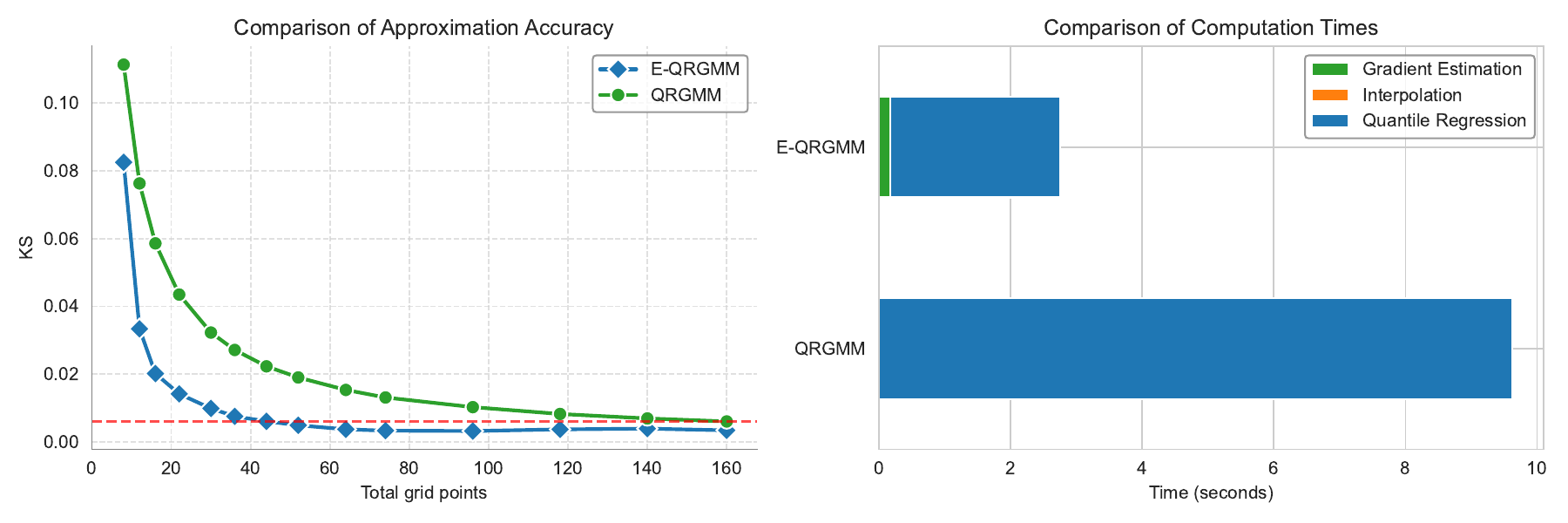}
    \caption{
Comparison of E-QRGMM and QRGMM under a normal test distribution with \( n = 10^5 \). 
\textbf{Left:}  E-QRGMM achieves the same accuracy with fewer grid points. 
\textbf{Right:} Compared to the \( n = 10^4 \) case, gradient estimation takes up an even smaller portion.
}

    \label{fig:n100000 E-QRGMM and QRGMM}
\end{figure}

When $n$ and the associated $m$ increase, the acceleration effect of the \text{E-QRGMM} becomes more pronounced. We therefore compare E-QRGMM and QRGMM on the normal test distribution with a larger dataset size $n=10^5$, keeping all other parameters the same. As shown in Figure~\ref{fig:n100000 E-QRGMM and QRGMM}, as $n$ grows, the proportion of runtime spent on gradient estimation further decreases, while the overall speedup of E-QRGMM over QRGMM increases due to the larger grid size. Overall, these results indicate that E-QRGMM scales favorably: it achieves comparable distributional accuracy with far fewer grid points, and the added cost of gradient estimation and cubic Hermite interpolation accounts for only a small fraction of the total runtime. Moreover, the speedup over QRGMM becomes more pronounced as the dataset size increases.

\begin{table}[!t]
\centering
\caption{Comparison of KS, WD, and per-model training time (mean $\pm$ standard error) across distributions and models. Time comparisons are conducted under equal CPU usage.}
\label{tab:performance}
\setlength{\tabcolsep}{6pt}
\renewcommand{\arraystretch}{1.1}
\resizebox{\textwidth}{!}{
\begin{tabular}{l*{9}{c}}
\toprule
 & \multicolumn{3}{c}{Normal} & \multicolumn{3}{c}{Halfnormal} & \multicolumn{3}{c}{Student's t} \\
\cmidrule(lr){2-4} \cmidrule(lr){5-7} \cmidrule(lr){8-10}
Models & KS & WD & Time(s) & KS & WD & Time(s) & KS & WD & Time(s) \\
\midrule

\multirow{2}{*}{GAN}
& 0.5064 & 1.5522 & 12.957
& 0.7067 & 1.8117 & 12.990
& 0.4098 & 1.1811 & 12.843 \\
& {\scriptsize$(\pm 2.29\text{e-}2)$} & {\scriptsize$(\pm 1.15\text{e-}1)$} & {\scriptsize$(\pm 3.40\text{e-}2)$}
& {\scriptsize$(\pm 2.60\text{e-}2)$} & {\scriptsize$(\pm 1.42\text{e-}1)$} & {\scriptsize$(\pm 3.63\text{e-}2)$}
& {\scriptsize$(\pm 2.12\text{e-}2)$} & {\scriptsize$(\pm 6.51\text{e-}2)$} & {\scriptsize$(\pm 3.73\text{e-}2)$} \\
\addlinespace[3pt]

\multirow{2}{*}{DDIM}
& 0.0534 & 0.1672 & 18.646
& 0.0953 & 0.1623 & 18.660
& 0.0465 & 0.1911 & 18.545 \\
& {\scriptsize$(\pm 3.30\text{e-}3)$} & {\scriptsize$(\pm 1.10\text{e-}2)$} & {\scriptsize$(\pm 2.50\text{e-}2)$}
& {\scriptsize$(\pm 3.39\text{e-}3)$} & {\scriptsize$(\pm 8.78\text{e-}3)$} & {\scriptsize$(\pm 2.33\text{e-}2)$}
& {\scriptsize$(\pm 2.96\text{e-}3)$} & {\scriptsize$(\pm 1.13\text{e-}2)$} & {\scriptsize$(\pm 2.55\text{e-}2)$} \\
\addlinespace[3pt]

\multirow{2}{*}{RectFlow}
& 0.0436 & 0.1270 & 18.427
& 0.0856 & 0.1313 & 18.428
& 0.0370 & 0.1675 & 18.394 \\
& {\scriptsize$(\pm 2.45\text{e-}3)$} & {\scriptsize$(\pm 7.43\text{e-}3)$} & {\scriptsize$(\pm 2.40\text{e-}2)$}
& {\scriptsize$(\pm 1.97\text{e-}3)$} & {\scriptsize$(\pm 4.28\text{e-}3)$} & {\scriptsize$(\pm 2.19\text{e-}2)$}
& {\scriptsize$(\pm 2.26\text{e-}3)$} & {\scriptsize$(\pm 7.09\text{e-}3)$} & {\scriptsize$(\pm 5.51\text{e-}2)$} \\
\addlinespace[3pt]

\multirow{2}{*}{E-QRGMM}
& \textbf{0.0110} & \textbf{0.0281} & \textbf{0.0960}
& \textbf{0.0113} & \textbf{0.0148} & \textbf{0.0879}
& \textbf{0.0110} & \textbf{0.0528} & \textbf{0.0990} \\
& {\scriptsize$(\pm 1.30\text{e-}4)$} & {\scriptsize$(\pm 6.90\text{e-}4)$} & {\scriptsize$(\pm 6.30\text{e-}5)$}
& {\scriptsize$(\pm 1.70\text{e-}4)$} & {\scriptsize$(\pm 5.00\text{e-}4)$} & {\scriptsize$(\pm 5.30\text{e-}5)$}
& {\scriptsize$(\pm 1.20\text{e-}4)$} & {\scriptsize$(\pm 8.90\text{e-}4)$} & {\scriptsize$(\pm 7.70\text{e-}5)$} \\

\bottomrule
\end{tabular}
}
\end{table}

\subsubsection{Accuracy and efficiency in distribution approximation}

We next compare E-QRGMM with several representative deep conditional generative baselines in terms of both distributional fidelity and computational efficiency. As reported in Table~\ref{tab:performance}, our method outperforms other deep learning models in training efficiency and distributional estimation accuracy measured by KS and WD.

\subsubsection{Statistical validation of bootstrap confidence interval}

The accuracy and training efficiency of E-QRGMM make it well-suited for bootstrap retraining, thereby facilitating the construction of covariate-dependent confidence intervals. We compare E-QRGMM with RectFlow using Algorithm~\ref{alg:bootstrap_general}. RectFlow is chosen as it achieves the best distributional accuracy among GAN, DDIM, and RectFlow. As shown in Table~\ref{tab:sta}, E-QRGMM produces shorter intervals and outperforms RectFlow. RectFlow attains empirical coverage close to \(1.00\) in most cases, which indicates overly conservative and thus uninformative intervals. This behavior is consistent with its larger distribution fitting error relative to E-QRGMM: under the same bootstrap procedure, poorer fit leads to greater estimation variability, yielding a broader empirical sampling distribution and consequently wider confidence intervals. Furthermore, our E-QRGMM method demonstrates robustness across different datasets and estimands of interest, which offers promising applicability for decision-making in high-risk scenarios.

\begin{table}[!t]
\centering
\caption{Empirical coverage and width (mean $\pm$ standard error) of the confidence interval constructed by \text{E-QRGMM} and RectFlow across different distribution types and estimands of interest with $B=100,$ $\alpha=0.1$.}
\resizebox{0.9\textwidth}{!}{
\begin{tabular}{llcccccc}
\toprule
 &  & \multicolumn{2}{c}{Normal} & \multicolumn{2}{c}{Halfnormal} & \multicolumn{2}{c}{Student's t} \\
\cmidrule(lr){3-4} \cmidrule(lr){5-6} \cmidrule(lr){7-8}
 Estimand & Model & Coverage & Width & Coverage & Width & Coverage & Width \\
\midrule

\multirow{4}{*}{Mean}
& \multirow{2}{*}{E-QRGMM}
  & 0.89 & \textbf{0.0481} & 0.90 & \textbf{0.0296} & 0.89 & \textbf{0.0612} \\
& & {\scriptsize$(\pm 3.13\text{e}-2)$} & {\scriptsize$(\pm 4.54\text{e}-4)$}
  & {\scriptsize$(\pm 3.00\text{e}-2)$} & {\scriptsize$(\pm 2.28\text{e}-4)$}
  & {\scriptsize$(\pm 3.13\text{e}-2)$} & {\scriptsize$(\pm 4.38\text{e}-4)$} \\
& \multirow{2}{*}{RectFlow}
  & 1.00 & 0.4187 & 1.00 & 0.3396 & 1.00 & 0.4495 \\
& & {\scriptsize$(\pm 0)$} & {\scriptsize$(\pm 3.64\text{e}-3)$}
  & {\scriptsize$(\pm 0)$} & {\scriptsize$(\pm 3.09\text{e}-3)$}
  & {\scriptsize$(\pm 0)$} & {\scriptsize$(\pm 3.80\text{e}-3)$} \\
\midrule

\multirow{4}{*}{Quantile}
& \multirow{2}{*}{E-QRGMM}
  & 0.90 & \textbf{0.0695} & 0.90 & \textbf{0.0551} & 0.91 & \textbf{0.0830} \\
& & {\scriptsize$(\pm 3.00\text{e}-2)$} & {\scriptsize$(\pm 7.24\text{e}-4)$}
  & {\scriptsize$(\pm 3.00\text{e}-2)$} & {\scriptsize$(\pm 5.56\text{e}-4)$}
  & {\scriptsize$(\pm 2.86\text{e}-2)$} & {\scriptsize$(\pm 9.07\text{e}-4)$} \\
& \multirow{2}{*}{RectFlow}
  & 1.00 & 0.4598 & 0.95 & 0.3779 & 1.00 & 0.5015 \\
& & {\scriptsize$(\pm 0)$} & {\scriptsize$(\pm 4.00\text{e}-3)$}
  & {\scriptsize$(\pm 2.18\text{e}-2)$} & {\scriptsize$(\pm 3.18\text{e}-3)$}
  & {\scriptsize$(\pm 0)$} & {\scriptsize$(\pm 4.27\text{e}-3)$} \\
\midrule

\multirow{4}{*}{Survival Func.}
& \multirow{2}{*}{E-QRGMM}
  & 0.90 & \textbf{0.0144} & 0.90 & \textbf{0.0143} & 0.91 & \textbf{0.0146} \\
& & {\scriptsize$(\pm 3.00\text{e}-2)$} & {\scriptsize$(\pm 1.41\text{e}-4)$}
  & {\scriptsize$(\pm 3.00\text{e}-2)$} & {\scriptsize$(\pm 1.38\text{e}-4)$}
  & {\scriptsize$(\pm 2.86\text{e}-2)$} & {\scriptsize$(\pm 1.38\text{e}-4)$} \\
& \multirow{2}{*}{RectFlow}
  & 1.00 & 0.0985 & 0.95 & 0.1062 & 1.00 & 0.0941 \\
& & {\scriptsize$(\pm 0)$} & {\scriptsize$(\pm 9.80\text{e}-4)$}
  & {\scriptsize$(\pm 2.18\text{e}-2)$} & {\scriptsize$(\pm 9.66\text{e}-4)$}
  & {\scriptsize$(\pm 0)$} & {\scriptsize$(\pm 8.25\text{e}-4)$} \\

\bottomrule
\end{tabular}
}
\label{tab:sta}
\end{table}

\subsection{Practically motivated experiments}\label{subsec: practical data}
The \((s,S)\) inventory problem is often solved by simulation because simple closed-form analytical formulas are typically unavailable in realistic settings, as real systems rarely satisfy the restrictive assumptions required for tractable analytical derivations \citep{pichitlamken2003combined,hong2006discrete,li2025additive}. Specifically, under a given \((s,S)\) policy, the cost is highly nonlinear and often lacks a tractable explicit form because demand and lead times are stochastic. Discrete-event simulation can mimic these  detailed stochastic dynamics and output the average cost under a specified \((s,S)\) policy, making it a natural tool for evaluating and optimizing \((s,S)\) policies.

We consider a simulator whose input is the covariate vector $\mathbf{x} = (1,s,S,\mu)^\top$ and whose output $Y$ is the  average cost. The decision variables are reorder point  $s \in [270,340]$ and the order-up-to level $S \in [380,450]$, and customer demand in each period is exponentially distributed with mean $\mu \in [310,340]$.  To construct the dataset, we uniformly draw $n=10^4$ covariate points from the domain of $\bx$ and run the simulator once at each point to obtain the corresponding outcomes. To access the ground truth for estimands at a specific covariate \(\bx^*\), we repeatedly run the simulator at \(\bx^*\) for \(10^5\) independent replications to obtain multiple realizations of \(Y(\bx^*)\). More details on the simulator construction, dataset generation, and evaluation protocol are provided in~\ref{app: inventory}. For all the experiments in Section \ref{subsec: practical data}, we use a training set $\{(\bx_i,y_i)\}_{i=1}^n$ with $n=10^4$ to fit generative models, and generate $K=10^5$ observations with covariate $ \bx^* = (1,320,420,330)^\top$ from every model to evaluate performance. All results are presented as mean values with standard errors derived from $N=100$ independent experimental replications. Implementation details for all models are provided in~\ref{app:detailed exp}.

On the practical dataset, Assumption~\ref{assumption:linear} no longer holds. However, by enriching E-QRGMM with nonlinear basis expansions, our method still performs well empirically. Specifically, we employ the quantile regression model \(\hat{Q}(\tau \mid \bx)=\bb(\bx)^\top\hat{\bbeta}(\tau)\), where \(\bx=(1,s,S,\mu)^\top\) and \(\bb(\bx)\) is the vector of basis functions
\[
\bb(\bx) = \bigl(1,\ s+S,\ S-s,\ 1/(S-s),\ \mu\bigr)^\top.
\]
This choice of basis functions is motivated by domain knowledge, capturing the nonlinear dependency of the average cost on the order quantity. Table~\ref{tab:performance inventory} summarizes the distributional accuracy of different conditional generative models. Overall, \text{E-QRGMM} attains the lowest KS and WD among all competitors while still requiring substantially less training time, suggesting that the basis-enriched quantile regression remains effective even when Assumption~\ref{assumption:linear} is violated in this practically motivated setting.

Beyond distributional fit, we further evaluate uncertainty quantification for estimands of interest at $\bx^*$. Table~\ref{tab:sta inventory} summarizes the empirical coverage and interval width of the resulting confidence intervals. Across all three estimands of interest, \text{E-QRGMM} attains coverage close to the nominal level $1-\alpha=0.9$ with markedly narrower intervals than RectFlow, suggesting that our method provides informative yet well-calibrated uncertainty quantification on the inventory management dataset.

\begin{table}[!t]
\centering
\footnotesize
\caption{Comparison of KS, WD, and per-model training time (mean $\pm$ standard error) on inventory management dataset. Time comparisons are conducted under equal CPU usage.}
\label{tab:performance inventory}
\setlength{\tabcolsep}{6pt}
\renewcommand{\arraystretch}{1.1}
\begin{tabular}{lccc}
\toprule
Model & KS & WD & Time(s) \\
\midrule

\multirow{2}{*}{GAN}
& 0.2018 & 4.0173 & 13.006 \\
& {\scriptsize$(\pm 1.09\text{e-}2)$} & {\scriptsize$(\pm 1.80\text{e-}1)$} & {\scriptsize$(\pm 3.25\text{e-}2)$} \\
\addlinespace[3pt]

\multirow{2}{*}{DDIM}
& 0.0428 & 1.0598 & 18.635 \\
& {\scriptsize$(\pm 2.73\text{e-}3)$} & {\scriptsize$(\pm 7.48\text{e-}2)$} & {\scriptsize$(\pm 2.49\text{e-}2)$} \\
\addlinespace[3pt]

\multirow{2}{*}{RectFlow}
& 0.0293 & 0.6676 & 18.444 \\
& {\scriptsize$(\pm 1.37\text{e-}3)$} & {\scriptsize$(\pm 3.37\text{e-}2)$} & {\scriptsize$(\pm 2.29\text{e-}2)$} \\
\addlinespace[3pt]

\multirow{2}{*}{E-QRGMM}
& \textbf{0.0132} & \textbf{0.2359} & \textbf{0.1219} \\
& {\scriptsize$(\pm 3.94\text{e-}4)$} & {\scriptsize$(\pm 7.79\text{e-}3)$} & {\scriptsize$(\pm 2.98\text{e-}4)$} \\

\bottomrule
\end{tabular}
\end{table}

\begin{table}[!t]
\centering
\footnotesize
\caption{Empirical coverage and width (mean $\pm$ standard error) of the confidence interval constructed by \text{E-QRGMM} and RectFlow across estimands of interest on inventory management dataset with $B=100,$ $\alpha=0.1$.}
\label{tab:sta inventory}
\setlength{\tabcolsep}{6pt}
\renewcommand{\arraystretch}{1.1}
\begin{tabular}{llcc}
\toprule
Estimand & Model & Coverage & Width \\
\midrule

\multirow{4}{*}{Mean}
& \multirow{2}{*}{E-QRGMM} & 0.90 & \textbf{0.5505} \\
&  & {\scriptsize$(\pm 3.00\text{e-}2)$} & {\scriptsize$(\pm 4.81\text{e-}3)$} \\
& \multirow{2}{*}{RectFlow} & 1.00 & 2.2306 \\
&  & {\scriptsize$(\pm 0)$} & {\scriptsize$(\pm 2.10\text{e-}2)$} \\
\midrule

\multirow{4}{*}{Quantile}
& \multirow{2}{*}{E-QRGMM} & 0.89 & \textbf{0.7992} \\
&  & {\scriptsize$(\pm 3.13\text{e-}2)$} & {\scriptsize$(\pm 1.04\text{e-}2)$} \\
& \multirow{2}{*}{RectFlow} & 1.00 & 2.4798 \\
&  & {\scriptsize$(\pm 0)$} & {\scriptsize$(\pm 2.47\text{e-}2)$} \\
\midrule

\multirow{4}{*}{Survival Func.}
& \multirow{2}{*}{E-QRGMM} & 0.90 & \textbf{0.0214} \\
&  & {\scriptsize$(\pm 3.00\text{e-}2)$} & {\scriptsize$(\pm 2.50\text{e-}4)$} \\
& \multirow{2}{*}{RectFlow} & 1.00 & 0.0658 \\
&  & {\scriptsize$(\pm 0)$} & {\scriptsize$(\pm 6.90\text{e-}4)$} \\

\bottomrule
\end{tabular}
\end{table}

\section{Conclusion and discussion}\label{conclusion}

We propose E-QRGMM, a computationally efficient generative metamodeling framework for conditional uncertainty quantification in simulation-based decision problems. By leveraging the gradient information of quantile functions and employing cubic Hermite interpolation, E-QRGMM accurately approximates the full conditional distribution while significantly reducing the number of required quantile regressions. This reduction enables bootstrap-based confidence interval construction with at least 100 cycles of generative model retraining, a scale that was previously impractical for simulation metamodels. Theoretically, E-QRGMM retains the $O_{\mathbb{P}}(n^{-1/2})$ convergence rate of QRGMM, while reducing quantile grid complexity from $O(n^{1/2})$ to $O(n^{1/5})$ in the central region. Empirically, E-QRGMM demonstrates a superior balance between distributional fidelity and computational efficiency compared to both QRGMM and other generative models, across a range of synthetic and practical datasets. In particular, E-QRGMM performs well in covariate-dependent uncertainty quantification for inventory applications, producing statistically valid and informative confidence intervals.

Our work paves the way for reliable and efficient conditional uncertainty quantification in simulation-driven decision-making. A promising future direction is extending E-QRGMM to handle multivariate outputs, where inverse distribution modeling becomes more challenging. 

\bibliographystyle{elsarticle-harv} 
\bibliography{references.bib}

\newpage
\appendix

\section{Quantile-regression-based Generative Metamodeling} \label{appendix:QRGMM}

The Quantile-Regression-Based Generative Metamodeling (QRGMM) method adopts the
OSOA (\emph{offline-simulation-online-application}) framework proposed by \citet{hong2019offline}, with the goal of constructing a fast and accurate generative surrogate for complex stochastic simulators. QRGMM enables conditional observation generation by learning an approximation to the conditional quantile function via quantile regression, followed by inverse transform sampling using interpolation. The algorithm proceeds in two stages:
\begin{algorithm}[H]
	\caption{Quantile-Regression-Based Generative Metamodeling}\label{alg:QRGMM}
	\begin{algorithmic}
		\State \textbf{Offline Stage:}
		\begin{itemize}[noitemsep,topsep=0pt,leftmargin=*]
			\item Input: Dataset \( \{(\bx_i, y_i)\}_{i=1}^n \), quantile regression model \( Q(\tau \mid \bx) \), integer \( m \).
			\item Create an equally-spaced grid \( \{\tau_j = j/m : j=1,\ldots,m-1\} \).
			\item Fit $Q(\tau_j \mid \bx)$ for each $\tau_j$ to obtain estimated quantile functions $\hat{Q}(\tau_j \mid \bx)$.
			\item Construct the quantile function \( \hat{Q}_O(\tau \mid \bx) \) via linear interpolation of \( \{(\tau_j, \hat{Q}(\tau_j \mid \bx))\}_{j=1}^{m-1} \):
\[
\hat{Q}_O(\tau \mid \bx) =
\begin{cases}
\hat{Q}(\tau_1 \mid \bx), & \tau < \tau_1, \\
\hat{Q}(\tau_j \mid \bx) + m (\tau - \tau_j)\bigl[\hat{Q}(\tau_{j+1} \mid \bx) - \hat{Q}(\tau_j \mid \bx)\bigr],
& \tau \in [\tau_j, \tau_{j+1}), \\
\hat{Q}(\tau_{m-1} \mid \bx), & \tau \ge \tau_{m-1}.
\end{cases}
\]
		\end{itemize}
		\State \textbf{Online Stage:}
		\begin{itemize}[noitemsep,topsep=0pt,leftmargin=*]
			\item Input: Covariate \( \bx = \bx^* \), number of observations \( K \).
			\item Generate \( \{u_k\}_{k=1}^K \sim \mathcal{U}(0,1) \).
			\item Output: Observations \( \{\hat{Y}_k(\bx^*) = \hat{Q}_O(u_k \mid \bx^*) : k=1,\ldots,K\} \).
		\end{itemize}
	\end{algorithmic}
\end{algorithm}
\paragraph{Generation efficiency} The key advantage of QRGMM is its real-time observation generation speed. Once the offline quantile regressions are trained, the online stage only involves inverse sampling, which is computationally trivial. As reported by \citet{hong2023learning}, QRGMM can generate $10^5$ conditional observations in under $0.01$ seconds, making it suitable for real-time decision-making tasks.

\paragraph{Theoretical guarantees} QRGMM also enjoys strong theoretical properties. \citet{hong2023learning} rigorously establish that the generative observations produced by QRGMM converge in distribution to the true conditional distribution, under standard regularity conditions. They further characterize the convergence rate with respect to both the number of grid points $m$ and the number of training data $n$, supporting the choice of $m$ in practice.

\paragraph{Accuracy and stability} Compared with alternative generative models such as Conditional Wasserstein GANs \citep{gulrajani2017improved}, QRGMM achieves substantially higher accuracy and greater stability in learning conditional distributions. As demonstrated in \citet{hong2023learning}, QRGMM avoids common pitfalls such as mode collapse and unstable training dynamics, which frequently affect GAN-based methods. Furthermore, QRGMM requires tuning only a single hyperparameter---the number of quantile grid points $m$---whose theoretical choice is well-supported; it is recommended to set $m = O(\sqrt{n})$ based on convergence rate analysis. This minimal tuning requirement enhances the method's ease of use and makes it particularly suitable for practical applications where robustness and reliability are critical.

The combination of the aforementioned advantages renders QRGMM a highly appealing generative metamodel for simulation-based inference involving covariates.

\section{Gradient estimation}\label{appendix:gradient_proof}

Before deriving the Proposition~\ref{pro:gradient_estimation}, we first state the required lemma from \citet{hong2010pathwise}.
\begin{lemma}[\citet{hong2010pathwise}]\label{lemma:hong2010}
For $L(\tau)$ satisfying $\frac{d}{d\tau}L(\tau)$ exists with probability $1$ (w.p.$1$) in its domain and there exists a random variable $\mathcal{R}$ such that $\mathbb{E}[\mathcal{R}]<\infty$ and $|L(\tau+h)-L(\tau)|\le\mathcal{R}|h|$ for any $h$ that is close enough to $0$. Denote $F(t,\tau) = \mathbb{P}\{L(\tau)\le t\} = \mathbb{E}[\mathbbm{1}\{L(\tau)\le t \}]$. 
 If for any $\tau$ in the domain, $F(t,\tau)$ is $\mathbb{L}^1$ continuous at $(t,\tau)$, then
\[
\frac{d}{d\tau} F(t,\tau) = -\frac{d}{dt}\left\{\mathbb{E}\left[\mathbbm{1}\{L(\tau)\le t\}\cdot\frac{d}{d\tau}L(\tau)\right]\right\}.
\]
\end{lemma}

\subsection*{Derivation of Proposition~\ref{pro:gradient_estimation}}
First we restate the 
Recall that quantile regression solves the population form of Eq.~\eqref{pinball loss}
\[
{\bbeta}(\tau) \equiv \mathop{\arg\min}\limits_{\beta \in \mathbb{R}^p} \, \mathbb{E}\left[ \rho_\tau(Y -  \bx^\top \bbeta)\right].
\]
Let $f(\bbeta)=\mathbb{E}\left[\rho_{\tau}\left(Y-\bx^{\top} \bbeta\right)\right]$. Then by the first-order optimality condition we have
\[
\frac{d}{d\bbeta} f(\bbeta) = - \mathbb{E}\left[(\tau - \mathbbm{1}\{Y-\bx^\top\bbeta<0\})\cdot \bx\right] = 0.
\]
$\bbeta(\tau)$ at quantile level $\tau$ satisfies
\begin{equation}\label{eq:first-order condition}
     \tau \cdot \mathbb{E}[\bx]=\mathbb{E}\left[\bx \cdot \mathbbm{1}\left\{Y-\bx^\top \bbeta(\tau)<0\right\}\right].
\end{equation}
Under Assumptions~\ref{ass:pathwise} and \ref{ass:residual_smoothness}, the functions $L(\tau) = Y-\bx^\top\bbeta(\tau)$ and $F(t,\tau) = \mathbb{E}\left[\mathbbm{1}\{L(\tau)\le t\}\right]$ satisfy the regularity conditions stated in Lemma~\ref{lemma:hong2010}. Note that $\bx$ has a compact domain $\mathcal{X}\subset\mathbb{R}^p$,  and is independent of both $t$ and $\tau$. Then, by Lemma~\ref{lemma:hong2010} and finite difference,
\[
\begin{aligned}
\frac{d}{d \tau}\,\mathbb{E}\!\left[\bx \cdot \mathbbm{1}\!\left\{Y-\bx^\top \bbeta(\tau)<0\right\}\right]
&=\left.\frac{d}{dt}\,\mathbb{E}\!\left[\bx \bx^\top \mathbbm{1}\!\left\{Y-\bx^\top \bbeta(\tau)<t\right\}\right]\right|_{t=0}
\cdot \frac{d}{d \tau}\,\bbeta(\tau) \\
&=\lim_{\delta \rightarrow 0^+}
\frac{1}{2\delta}\,\mathbb{E}\!\left[\bx \bx^\top \mathbbm{1}\!\left\{-\delta<Y-\bx^\top \bbeta(\tau)<\delta\right\}\right]
\cdot \frac{d}{d \tau}\,\bbeta(\tau).
\end{aligned}
\]
Let $\Lambda(\tau)=\lim _{\delta \rightarrow 0^+} \frac{1}{2 \delta} \mathbb{E}\left[\bx \bx^\top \mathbbm{1}\left\{-\delta<Y-\bx^\top \bbeta(\tau)<\delta\right\}\right]$. Then, by differentiating w.r.t. $\tau$ on both side of Eq.~\eqref{eq:first-order condition}, we have
\[
\mathbb{E}[\bx]=\Lambda(\tau) \cdot \frac{d}{d \tau} \bbeta(\tau) \Rightarrow \frac{d}{d \tau} \bbeta(\tau)=\Lambda^{-1}(\tau) \cdot \mathbb{E}[\bx].
\]
And thus,
\[
D(\tau \mid \mathbf{x}) = \mathbf{x}^{\top} \frac{d}{d\tau} \bbeta(\tau) = \mathbf{x}^{\top} \Lambda^{-1}(\tau) \mathbb{E}[\bx], 
\]
which concludes the proof. \hfill $\square$

When there are data $\{(\bx_i,y_i)\}_{i=1}^n$, we can estimate $\mathbb{E}[\bx]$ by $\bar{\bx} = \frac{1}{n}\sum_{i=1}^n \bx_i$. For $\Lambda(\tau)$, the estimator is 
\[
   \hat{\Lambda}(\tau) = \frac{1}{2 \delta_n} \frac{1}{n} \sum_{i=1}^n \bx_i \bx_i^\top \cdot \mathbbm{1}\{-\delta_n < y_i - \bx_i^\top \hat{\bbeta}(\tau) < \delta_n\},
\]
where \( \hat{\bbeta}(\tau) \) is the sample quantile regression defined by Eq.~\eqref{pinball loss} with the parameter $\delta_n = O(n^{-1/5})$ as suggested in \citet{hong2010pathwise}.

\section{Instability of tail gradient estimation}\label{app:tail gradient}

\begin{figure}[!htbp]
    \centering
    \includegraphics[width=0.7\linewidth]{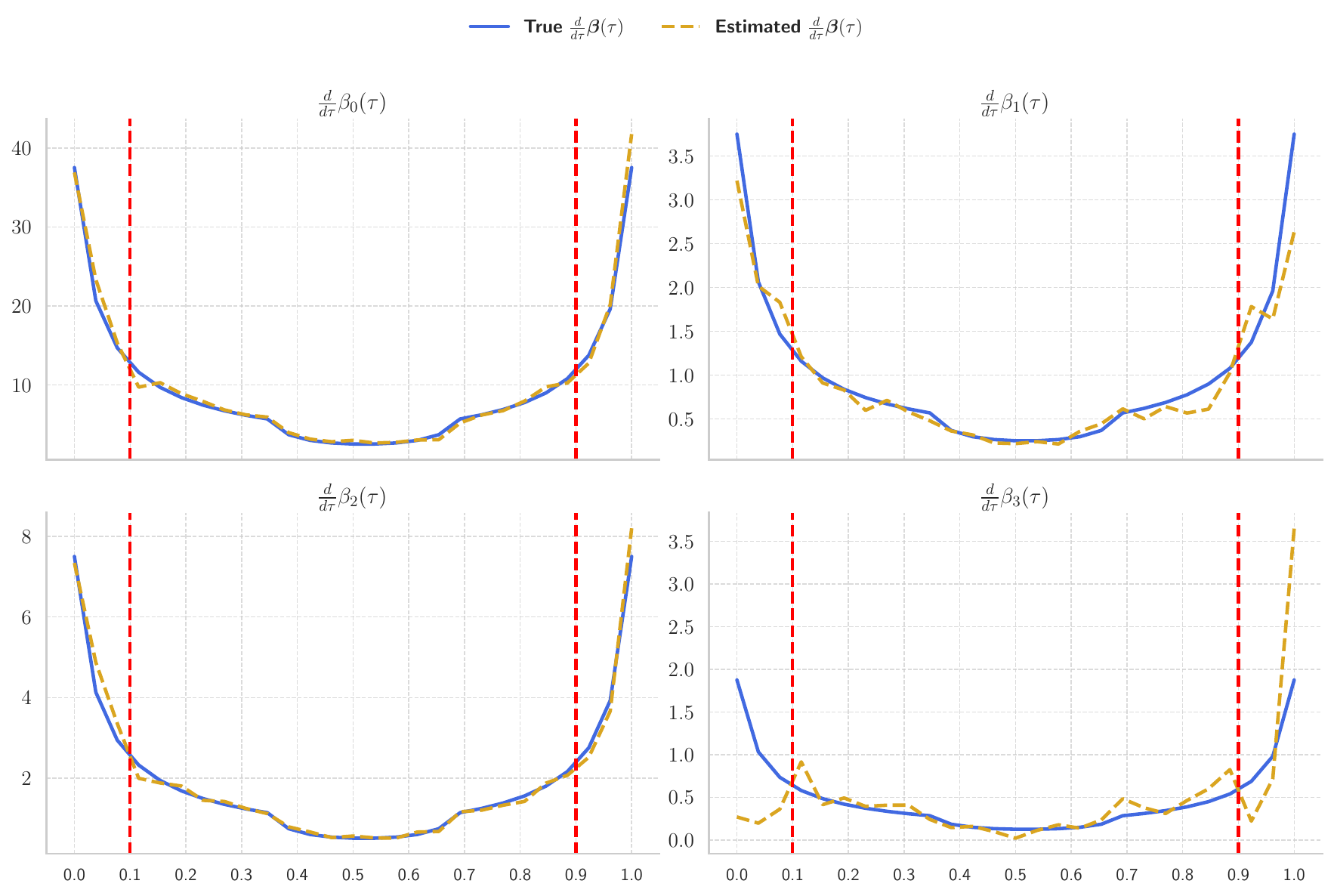}
    \caption{
Estimated versus true gradient curves of $\frac{d}{d\tau}\bbeta(\tau)$ under the normal test distribution, . The blue lines represent the true gradients, while the  yellow lines show the estimated gradients. Vertical red  lines at $\tau=0.1$ and $\tau=0.9$ mark the tail regions. Notably, the estimated gradients closely match the ground truth in the central region, but exhibit significant numerical inaccuracies near the distributional tails.
}

    \label{fig:tail gradient estimation}
\end{figure}

We illustrate the gradient estimation behavior under the normal test distribution defined in Section~\ref{empirical} with $n=10^4$, grid point design $\{j/m:j=1,\ldots,m-1\},m=\sqrt{n}=100$. As shown in Figure~\ref{fig:tail gradient estimation}, gradient estimation in the tail region suffers from substantial numerical errors.

Beyond the graphical illustrations, we further examine the numerical stability. Denote the gradient of \(\boldsymbol{\beta}(\tau)\) at quantile level \(\tau\) as \(D(\tau)\). For the normal distribution case described in Section~\ref{empirical}, we calculate the relative \(L_1\) errors \(\|\hat{D}(\tau)-D(\tau)\|_1/\|D(\tau)\|_1\) under different dataset sizes for quantile levels \([0.01,0.05,0.1,0.2,0.3,0.5,0.7,0.8,0.9,0.95,0.99]\). We report the standard deviation of relative \(L_1\) error of \(D(\tau)\) within \(100\) repeated experiments.

\begin{table}[H]
\centering
\caption{Standard deviation of $L_1$ errors under different dataset sizes for various quantile levels.}
\label{tab:gradient sensitivity}
\resizebox{\textwidth}{!}{
\begin{tabular}{c|ccccccccccc}
\hline
$\tau$   & 0.01   & 0.05   & 0.1    & 0.2    & 0.3    & 0.5    & 0.7    & 0.8    & 0.9    & 0.95   & 0.99   \\
\hline
$n=10^3$ & 5.3995 & 1.0144 & 0.4827 & 0.2631 & 0.2926 & 0.2494 & 0.2766 & 0.3236 & 0.4594 & 0.6690 & 2.6767 \\
$n=10^4$ & 0.3805 & 0.1551 & 0.1068 & 0.0808 & 0.0933 & 0.0843 & 0.0919 & 0.0779 & 0.1114 & 0.1627 & 0.4022 \\
$n=10^5$ & 0.0821 & 0.0482 & 0.0374 & 0.0317 & 0.0240 & 0.0236 & 0.0270 & 0.0249 & 0.0347 & 0.0427 & 0.0995 \\
\hline
\end{tabular}
}
\end{table}

A large standard deviation indicates that the estimated gradient is unstable and sensitive to sampling variability. As shown in Table~\ref{tab:gradient sensitivity}, the standard deviation of relative error becomes increasingly large in the extreme tails (close to $0$ or $1$) because of data sparsity, while it remains much smaller and more stable in the central quantile range. This highlights the importance of introducing truncation parameters $\tau_{\mathsf{L}}$ and $\tau_{\mathsf{U}}$ to mitigate instability in the numerical estimates.

In addition, the experimental results in Table~\ref{tab:gradient sensitivity} can also be viewed as a sensitivity analysis for the truncation parameters. The results show that the range of quantiles with stable gradients expands as the dataset size grows. For example, with $n=10^3$, quantiles roughly between $0.1$ and $0.9$ exhibit acceptable stability, whereas with $n=10^5$ or more, the error is well controlled even closer to the tails (e.g., $\tau=0.05$ or $0.95$). This suggests that the choice of truncation parameters $\tau_L$ and $\tau_U$ should depend on the data size: in smaller datasets, it is safer to use a narrower interval, while larger datasets allow more aggressive inclusion of extreme quantiles. 

\section{Supplementary materials for the theoretical analysis}

\subsection{Assumptions}\label{app:assumptions}

\begin{assumption}[Smoothness of the Quantile Function]\label{ass:smoothness}
	For any $\bx \in \mathcal{X}$, the conditional quantile function \( Q(\tau \mid \mathbf{x}) \) is four times continuously differentiable with respect to \( \tau \) over the interval \( \tau \in [\tau_{\mathsf{L}}, \tau_{\mathsf{U}}] \), where \( 0 < \tau_{\mathsf{L}} < \tau_{\mathsf{U}} < 1 \). Furthermore, its fourth derivative is uniformly bounded over the interval \( \tau \in [\tau_{\mathsf{L}}, \tau_{\mathsf{U}}] \), i.e., there exists a constant \( M' < \infty \) such that
	\[
	\sup_{\tau \in [\tau_{\mathsf{L}}, \tau_{\mathsf{U}}]} \left| Q^{(4)}(\tau \mid \mathbf{x}) \right| \le M'.
	\]
\end{assumption}
This condition is commonly imposed in the context of cubic Hermite interpolation, ensuring sufficient smoothness for accurate approximation of the quantile function. For general continuous random variables, this requirement is typically satisfied.

\begin{assumption}[Quantile Regression Conditions]\label{ass:quantile_regression}
We state below a set of conditions commonly used for quantile regression.
	\begin{enumerate}[label=(3.\alph*), left=0pt]
		\item \label{ass:cons2a} The conditional distribution function $F_Y(\cdot \mid \bx)$ is absolutely continuous with respect to the Lebesgue measure, and admits a conditional density $f_Y(\cdot \mid \bx)$, which shares a common support across all $\bx \in \mathcal{X}$.

		\item \label{ass:cons2b} Given any quantile interval \( 0 < \tau_{\mathsf{L}} < \tau_{\mathsf{U}} < 1 \), the conditional density function \( f_Y(\cdot \mid \bx) \) evaluated at the conditional quantile \( F_Y^{-1}(\tau \mid \bx)= Q(\tau \mid \mathbf{x})\) is uniformly bounded above and bounded away from zero across \( \tau \in [\tau_{\mathsf{L}}, \tau_{\mathsf{U}}] \) and \( \bx \in \mathcal{X} \), i.e.,
\[
\begin{aligned}
\inf_{\tau \in [\tau_{\mathsf{L}}, \tau_{\mathsf{U}}]}
\inf_{\bx \in \mathcal{X}}
f_Y\!\bigl(F_Y^{-1}(\tau \mid \bx)\mid \bx\bigr)
&> 0, \\
\sup_{\tau \in [\tau_{\mathsf{L}}, \tau_{\mathsf{U}}]}
\sup_{\bx \in \mathcal{X}}
f_Y\!\bigl(F_Y^{-1}(\tau \mid \bx)\mid \bx\bigr)
&< \infty.
\end{aligned}
\]

		\item \label{ass:cons3a} There exist positive definite matrices $D_0$ and $D_1(\tau)$ such that, for any $0 < \tau_{\mathsf{L}} < \tau_{\mathsf{U}} < 1$, the following limits hold almost surely as $n \to \infty$:
		\[
		\frac{1}{n} \sum_{i=1}^n \bx_i \bx_i^\top \to D_0,
		\quad
		\frac{1}{n} \sum_{i=1}^n f_Y(F_Y^{-1}(\tau \mid \bx_i) \mid \bx_i) \bx_i \bx_i^\top\to D_1(\tau),
		\]
		where the second convergence is uniform over $\tau \in [\tau_{\mathsf{L}}, \tau_{\mathsf{U}}]$.
		
		\item \label{ass:cons3b} The covariate space \( \mathcal{X} \subset \mathbb{R}^p \) is compact.
	\end{enumerate}
\end{assumption}
These conditions are standard in the quantile regression literature and are adopted from \citet{hong2023learning}. They ensure the well-posedness of the quantile regression framework by guaranteeing the existence and boundedness of conditional densities, as well as the asymptotic stability of the covariate structure. Such assumptions are critical for establishing the consistency and asymptotic normality of quantile regression estimators.

\begin{assumption}[Pathwise Derivative of Residuals]\label{ass:pathwise}
	Denote residuals as \( L(\tau) =  Y - \bx^\top \bbeta(\tau) \). 
	For any \(\tau \in [\tau_{\mathsf{L}}, \tau_{\mathsf{U}}]\), the derivative \( L'(\tau) = \frac{d}{d\tau} [Y - \bx^\top \bbeta(\tau)] = -\bx^\top \frac{d}{d\tau} \bbeta(\tau) \) exists with probability 1, and there exists a random variable \(\mathcal{R}\) with \(\mathbb{E}(\mathcal{R}) < \infty\) such that for any \(h\) sufficiently small,
	\[
	\left| [Y - \bx^\top \bbeta(\tau + h)] - [Y - \bx^\top \bbeta(\tau)] \right| \leq \mathcal{R} |h|.
	\]
\end{assumption}

\begin{assumption}[Smoothness of the Residual Distribution]\label{ass:residual_smoothness}
	For any \(\tau \in [\tau_{\mathsf{L}}, \tau_{\mathsf{U}}]\), the cumulative distribution function of the residual, \( F(t, \tau) =  \mathbb{P}\{Y - \bx^\top \bbeta(\tau) \leq t\} \), is continuously differentiable with respect to \( t \) at \( t = 0 \), and the density \( f(t, \tau) = \frac{\partial}{\partial t} F(t, \tau) \) satisfies \(0< f(0, \tau) < \infty \) over \([\tau_{\mathsf{L}}, \tau_{\mathsf{U}}]\), and for any $\bx\in \mathcal{X} $, the conditional density \(f(t,\tau\mid
\bx)\) is continuous and bounded in a neighborhood of $t=0$ over \([\tau_{\mathsf{L}}, \tau_{\mathsf{U}}]\).
\end{assumption}

\begin{assumption} [Differentiability of Moment Functions] \label{ass:moment_diff}
Define
$
h(t, \tau) = \mathbb{E}[L'(\tau) \cdot \mathbbm{1}_{\{L(\tau) \leq t\}}]$, and $h_\gamma(t, \tau) = \mathbb{E}[|L'(\tau)|^\gamma \cdot \mathbbm{1}_{\{L(\tau) \leq t\}}]
$. For any \(\tau \in [\tau_{\mathsf{L}}, \tau_{\mathsf{U}}]\) and \(t = 0\):  
\begin{itemize}
\item \(h_2(t, \tau) = \mathbb{E}[(L'(\tau))^2 \cdot \mathbbm{1}_{\{L(\tau) \leq t\}}]\) is differentiable with respect to \(t\), and \(h_2'(0, \tau) > 0\).  
\item For some \(\gamma > 0\), \(h_{2+\gamma}(t, \tau) = \mathbb{E}[|L'(\tau)|^{2+\gamma} \cdot \mathbbm{1}_{\{L(\tau) \leq t\}}]\) is differentiable with respect to \(t\).  
\item \(h(t, \tau)\) has a third derivative \(h^{(3)}(t, \tau)\) at \(t = 0\).
\end{itemize}
\end{assumption}
The preceding three assumptions are derived from \citet{hong2010pathwise}, where they facilitate the analysis of pathwise derivatives and residual distributions in quantile regression models. These conditions ensure the differentiability and boundedness of key moment functions, enabling precise characterization of the asymptotic behavior of estimators. A more detailed discussion of their implications and applications can be found in \citet{hong2010pathwise}.

\subsection{Proofs}\label{app:E-QRGMM proofs}

\subsubsection*{Proof of Lemma~\ref{lemma:interpolation_error}}

For any $j \in \{1, \ldots, m-1\}$ such that $[\tau_j, \tau_{j+1}] \cap [\tau_{\mathsf{L}} + \frac{1}{m}, \tau_{\mathsf{U}} - \frac{1}{m}] \neq \emptyset$, let $\tau \in [\tau_j, \tau_{j+1}] \cap [\tau_{\mathsf{L}} + \frac{1}{m}, \tau_{\mathsf{U}} - \frac{1}{m}]$.
By the standard cubic Hermite interpolation error bound (see \citet{de1978practical}), we have:
\[
\left| Q_C(\tau \mid \mathbf{x}^*) - Q(\tau \mid \mathbf{x}^*) \right| \leq \frac{1}{384} \cdot (\tau_{j+1} - \tau_j)^4 \cdot \sup_{\xi \in [\tau_j, \tau_{j+1}]} \left| Q^{(4)}(\xi \mid \mathbf{x}^*) \right|.
\]
Since the quantile grid is uniformly spaced with $\tau_{j+1} - \tau_j = \frac{1}{m}$, and Assumption~\ref{ass:smoothness} ensures that \\
$\sup_{\xi \in [\tau_j, \tau_{j+1}]} \left| Q^{(4)}(\xi \mid \mathbf{x}^*) \right| \leq M$ because $[\tau_j, \tau_{j+1}]\subset[\tau_{\mathsf{L}},\tau_{\mathsf{U}}]$ for such $j$, we obtain:
\[
\left| Q_C(\tau \mid \mathbf{x}^*) - Q(\tau \mid \mathbf{x}^*) \right| \leq \frac{M}{384} \cdot \frac{1}{m^4}.
\]
Finally, taking the supremum over \( \tau \in [\tau_{\mathsf{L}}+\frac{1}{m}, \tau_{\mathsf{U}}-\frac{1}{m}] \) across all subintervals, we have:
\[
\sup_{\tau \in [\tau_{\mathsf{L}} + \frac{1}{m}, \tau_{\mathsf{U}} - \frac{1}{m}]} \left| Q_C(\tau \mid \mathbf{x}^*) - Q(\tau \mid \mathbf{x}^*) \right| \leq \frac{M}{384} \cdot \frac{1}{m^4},
\]
which completes the proof. \hfill $\square$

\subsubsection*{Proof of Lemma~\ref{lemma:gradient_estimation_error}}

\textit{Step 1: Error Decomposition.} By proposition~\ref{pro:gradient_estimation}, the gradient estimation error is:
\[
\hat{D}(\tau_j \mid \mathbf{x}^*) - D(\tau_j \mid \mathbf{x}^*) = \mathbf{x}^{*^\top} \left( \hat{\Lambda}^{-1}(\tau_j) \bar{\bx} - \Lambda^{-1}(\tau_j) \mathbb{E}[\bx] \right).
\]
Decompose the inner term:
\[
\hat{\Lambda}^{-1}(\tau_j) \bar{\bx} - \Lambda^{-1}(\tau_j) \mathbb{E}[\bx] = (\hat{\Lambda}^{-1}(\tau_j) - \Lambda^{-1}(\tau_j)) \bar{\bx} + \Lambda^{-1}(\tau_j) (\bar{\bx} - \mathbb{E}[\bx]).
\]
The error becomes:
\[
\hat{D}(\tau_j \mid \mathbf{x}^*) - D(\tau_j \mid \mathbf{x}^*) = \mathbf{x}^{*^\top} \left[ (\hat{\Lambda}^{-1}(\tau_j) - \Lambda^{-1}(\tau_j)) \bar{\bx} + \Lambda^{-1}(\tau_j) (\bar{\bx} - \mathbb{E}[\bx]) \right].
\]
Since \(\mathbf{x}^*\) is fixed, we analyze the rates of \(\bar{\bx} - \mathbb{E}[\bx]\) and \(\hat{\Lambda}^{-1}(\tau_j) - \Lambda^{-1}(\tau_j)\).

\textit{Step 2: Rate of \(\bar{\bx} - \mathbb{E}[\bx]\).}
By the central limit theorem, under the moment condition:
\[
\bar{\bx} - \mathbb{E}[\bx] = O_p(n^{-1/2}),
\]
since Assumption \ref{ass:cons3a} implies that \(\mathbb{E}[\bx \bx^\top] < \infty\), this ensures finite variance.

\textit{Step 3: Decomposition of \(\hat{\Lambda}(\tau_j) - \Lambda(\tau_j)\).} Define an intermediate term:
\[
\bar{\Lambda}(\tau_j) = \frac{1}{2 \delta_n} \frac{1}{n} \sum_{i=1}^n \bx_i \bx_i^\top \mathbbm{1}\{-\delta_n < y_i - \bx_i^\top \bbeta(\tau_j) < \delta_n\},
\]
using the true \(\bbeta(\tau_j)\). Then:
\[
\hat{\Lambda}(\tau_j) - \Lambda(\tau_j) = \left(\hat{\Lambda}(\tau_j) - \bar{\Lambda}(\tau_j)\right) + \left(\bar{\Lambda}(\tau_j) - \Lambda(\tau_j)\right).
\]
We further analyze the rates of $\hat{\Lambda}(\tau_j) - \bar{\Lambda}(\tau_j)$ and $\bar{\Lambda}(\tau_j) - \Lambda(\tau_j)$.

\textit{Step 4: Rate of $\bar{\Lambda}(\tau_j) - \Lambda(\tau_j)$.} Theorem 3 from \citet{hong2010pathwise} states that under Assumptions \ref{ass:pathwise},\ref{ass:residual_smoothness}, and \ref{ass:moment_diff},
if \(n \delta_n^5 \to a < \infty\), then $\sqrt{2 n \delta_n} [\bar{\Lambda}(\tau_j) - \Lambda(\tau_j)]$ converge to a normal distribution. Thus, choose \(\delta_n = \tilde{c} n^{-1/5}\): We have
\[
 \bar{\Lambda}(\tau_j) - \Lambda(\tau_j) = O_p(n^{-2/5}).
\]

\textit{Step 5: Rate of $\hat{\Lambda}(\tau_j)  - \bar{\Lambda}(\tau_j) $.}
The difference is:  
\[
\frac{1}{2 \delta_n} \frac{1}{n} \sum_{i=1}^n \bx_i \bx_i^\top \left[ \mathbbm{1}\{-\delta_n < y_i - \bx_i^\top \hat{\bbeta}(\tau_j) < \delta_n\} - \mathbbm{1}\{-\delta_n < y_i - \bx_i^\top \bbeta(\tau_j) < \delta_n\} \right].
\]  
Define:  
\[
W_i = \frac{1}{2 \delta_n} \bx_i \bx_i^\top \left[ \mathbbm{1}\{-\delta_n < y_i - \bx_i^\top \hat{\bbeta}(\tau_j) < \delta_n\} - \mathbbm{1}\{-\delta_n < y_i - \bx_i^\top \bbeta(\tau_j) < \delta_n\} \right],
\]  
so that:  
\[
\hat{\Lambda}(\tau_j) - \bar{\Lambda}(\tau_j) = \frac{1}{n} \sum_{i=1}^n W_i.
\]  
For $\tau_j\in[\tau_{\mathsf{L}},\tau_{\mathsf{U}}]$, Let \(\Delta \bbeta(\tau_j)=\hat{\bbeta}(\tau_j) - \bbeta(\tau_j)\), then by proposition 3 of \citet{hong2023learning}, \(\Delta \bbeta(\tau_j)= O_p(n^{-1/2})\). And as \(\bx_i\) is bounded, the perturbation \(\bx_i^\top \Delta \bbeta(\tau_j) = O_p(n^{-1/2})\). Define: 
\[
\begin{aligned}
A_i &= \{-\delta_n < y_i - \bx_i^\top \bbeta(\tau_j) < \delta_n\},\\
\hat{A}_i &= \{-\delta_n < y_i - \bx_i^\top \hat{\bbeta}(\tau_j) < \delta_n\}= \{-\delta_n +\bx_i^\top\Delta \bbeta(\tau_j) < y_i - \bx_i^\top \bbeta(\tau_j) < \delta_n+\bx_i^\top\Delta \bbeta(\tau_j)\} .
\end{aligned}
\]
The difference \(\mathbbm{1}_{\hat{A}_i} - \mathbbm{1}_{A_i}\) arises from the perturbation \(\Delta \bbeta(\tau_j)\), so the event \(\mathbbm{1}_{\hat{A}_i} \neq \mathbbm{1}_{A_i}\) occurs when \(y_i - \bx_i^\top \bbeta(\tau_j)\) lies in boundary regions due to the shift \(\bx_i^\top \Delta \bbeta(\tau_j)\). If \(\bx_i^\top \Delta \bbeta(\tau_j) > 0\), \(\hat{A}_i\) shifts right, and the difference is non-zero in:
\begin{itemize}
\item \([-\delta_n, -\delta_n + \bx_i^\top \Delta \bbeta(\tau_j)]\) (when \(y_i - \bx_i^\top \bbeta(\tau_j)\) in \(A_i\), not \(\hat{A}_i\)),
\item \([\delta_n,\delta_n + \bx_i^\top \Delta \bbeta(\tau_j) ]\) (when \(y_i - \bx_i^\top \bbeta(\tau_j)\) in \(\hat{A}_i\), not \(A_i\)).
\end{itemize}
Each region has width \(|\bx_i^\top \Delta \bbeta(\tau_j)| = O_p(n^{-1/2})\). Given the residual \(y_i - \bx_i^\top \bbeta(\tau_j)\) has continuous and bounded conditional density \(f(t,\tau_j\mid
\bx_i)\) in a neighborhood of $t=0$ by Assumption~\ref{ass:residual_smoothness}, and since \(\delta_n = O(n^{-1/5})\) is small, approximate \(f(\pm \delta_n, \tau_j \mid \bx_i) \approx f(0, \tau_j \mid \bx_i)\). The probability for one region is:
\[
\mathbb{P}(y_i - \bx_i^\top \bbeta(\tau_j) \in [-\delta_n, -\delta_n + \bx_i^\top \Delta \bbeta(\tau_j)] \mid \bx_i, \hat{\bbeta}(\tau_j))  \approx f(0,\tau_j\mid\bx_i) \cdot |\bx_i^\top \Delta \bbeta(\tau_j)|.
\]
Total probability across both regions:
\[
\mathbb{P}(\mathbbm{1}_{\hat{A}_i} \neq \mathbbm{1}_{A_i} \mid \bx_i, \hat{\bbeta}(\tau_j)) \approx 2 f(0,\tau_j\mid\bx_i) |\bx_i^\top \Delta \bbeta(\tau_j)|.
\]
Since \(f(0,\tau_j\mid\bx_i)\) is bounded and \(|\bx_i^\top \Delta \bbeta(\tau_j)| = O_p(n^{-1/2})\):
\[
\mathbb{P}(\mathbbm{1}_{\hat{A}_i} \neq \mathbbm{1}_{A_i} \mid \bx_i, \hat{\bbeta}) = O_p(n^{-1/2}).
\] 
For the case that \(\bx_i^\top \Delta \bbeta(\tau_j) < 0\), the analysis is similar.\\
\textit{Conditional Expectation:} 
\[
\mathbb{E}[W_i \mid \bx_i, \hat{\bbeta}(\tau_j)] = \frac{1}{2 \delta_n} \bx_i \bx_i^\top \mathbb{E}\left[ \mathbbm{1}_{\hat{A}_i} - \mathbbm{1}_{A_i} \mid \bx_i, \hat{\bbeta}(\tau_j) \right].
\]  
Since \(\left| \mathbbm{1}_{\hat{A}_i} - \mathbbm{1}_{A_i} \right| \leq 1\) and occurs with probability \(O_p(n^{-1/2})\):  
\[
\left| \mathbb{E}[\mathbbm{1}_{\hat{A}_i} - \mathbbm{1}_{A_i} \mid \bx_i, \hat{\bbeta}(\tau_j)] \right| \leq \mathbb{P}(\mathbbm{1}_{\hat{A}_i} \neq \mathbbm{1}_{A_i} \mid \bx_i, \hat{\bbeta}(\tau_j)) = O_p(n^{-1/2}).
\]  
With \(\delta_n = O(n^{-1/5})\), and \(\bx_i \bx_i^\top\) bounded:  
\[
\mathbb{E}[W_i \mid \bx_i, \hat{\bbeta}(\tau_j)] = O_p\left( \frac{n^{-1/2}}{n^{-1/5}} \right) = O_p(n^{-3/10}).
\]
\textit{Conditional Variance:}
\[
\text{Var}(W_i \mid \bx_i, \hat{\bbeta}(\tau_j)) = \mathbb{E}[W_i^2 \mid \bx_i, \hat{\bbeta}(\tau_j)] - \left( \mathbb{E}[W_i \mid \bx_i, \hat{\bbeta}(\tau_j)] \right)^2.
\]  
Since \(\left( \mathbbm{1}_{\hat{A}_i} - \mathbbm{1}_{A_i} \right)^2 \leq 1\):  
\[
\mathbb{E}[W_i^2 \mid \bx_i, \hat{\bbeta}(\tau_j)] \leq \frac{1}{4 \delta_n^2} \| \bx_i \bx_i^\top \|^2 \mathbb{P}(\mathbbm{1}_{\hat{A}_i} \neq \mathbbm{1}_{A_i} \mid \bx_i, \hat{\bbeta}(\tau_j)) = O_p\left( \frac{n^{-1/2}}{n^{-2/5}} \right) = O_p(n^{-1/10}),
\]  
and \(\left( \mathbb{E}[W_i \mid \bx_i, \hat{\bbeta}(\tau_j)] \right)^2 = O_p(n^{-6/10})\), so:  
\[
\text{Var}(W_i \mid \bx_i, \hat{\bbeta}(\tau_j)) = O_p(n^{-1/10}).
\]
\textit{Total Error:}  
\[
\hat{\Lambda}(\tau_j) - \bar{\Lambda}(\tau_j) = \mathbb{E}[\hat{\Lambda}(\tau_j) - \bar{\Lambda}(\tau_j) \mid \hat{\bbeta}(\tau_j)] + \left( \hat{\Lambda}(\tau_j) - \bar{\Lambda}(\tau_j) - \mathbb{E}[\hat{\Lambda}(\tau_j) - \bar{\Lambda}(\tau_j) \mid \hat{\bbeta}(\tau_j)] \right).
\]  
The conditional expectation is:  
\[
\mathbb{E}[\hat{\Lambda}(\tau_j) - \bar{\Lambda}(\tau_j) \mid \hat{\bbeta}(\tau_j)] = \frac{1}{n} \sum_{i=1}^n \mathbb{E}[W_i \mid \bx_i, \hat{\bbeta}(\tau_j)] = O_p(n^{-3/10}).
\]  
The centered term’s variance is:  
\[
\text{Var}\left( \hat{\Lambda}(\tau_j) - \bar{\Lambda}(\tau_j) \mid \hat{\bbeta}(\tau_j) \right) = \frac{1}{n} \text{Var}(W_i \mid \bx_i, \hat{\bbeta}(\tau_j)) = O_p(n^{-1 - 1/10}) = O_p(n^{-11/10}).
\]  
By Chebyshev's inequality, we have:  
\[
\hat{\Lambda}(\tau_j) - \bar{\Lambda}(\tau_j) - \mathbb{E}[\hat{\Lambda}(\tau_j) - \bar{\Lambda}(\tau_j) \mid \hat{\bbeta}(\tau_j)] = O_p(n^{-11/20}).
\]  
Thus:  
\[
\hat{\Lambda}(\tau_j) - \bar{\Lambda}(\tau_j) = O_p(n^{-3/10}) + O_p(n^{-11/20}).
\]

\textit{Step 6: Rate of \(\hat{\Lambda}^{-1}(\tau_j) - \Lambda^{-1}(\tau_j)\).}  
From Steps 4 and 5:  
\[
\hat{\Lambda}(\tau_j) - \Lambda(\tau_j) = \left( \hat{\Lambda}(\tau_j) - \bar{\Lambda}(\tau_j) \right) + \left( \bar{\Lambda}(\tau_j) - \Lambda(\tau_j) \right) = O_p(n^{-3/10}) + O_p(n^{-11/20}) + O_p(n^{-2/5}).
\]  
Since \(-3/10 > -2/5> -11/20 \), the dominant term is \(O_p(n^{-3/10})\):  
\[
\hat{\Lambda}(\tau_j) - \Lambda(\tau_j) = O_p(n^{-3/10}).
\]  
As
\[
\Lambda(\tau_j) = \lim_{\delta \to 0} \frac{1}{2\delta} \mathbb{E}\left[ \bx \bx^\top \mathbbm{1}\{-\delta < Y - \bx^\top \bbeta(\tau) < \delta\} \right] = f(0, \tau_j) \mathbb{E}[\bx \bx^\top],
\]
by Assumptions \ref{ass:cons3a}	and \ref{ass:residual_smoothness}, \(\Lambda(\tau_j)\) is positive definite (hence invertible) and bounded, the matrix inverse perturbation \citep{stewart1990matrix} gives:  
\[
\hat{\Lambda}^{-1}(\tau_j) - \Lambda^{-1}(\tau_j) = -\Lambda^{-1}(\tau_j) \left( \hat{\Lambda}(\tau_j) - \Lambda(\tau_j) \right) \Lambda^{-1}(\tau_j) + O_p\left( \|\hat{\Lambda}(\tau_j) - \Lambda(\tau_j)\|^2 \right).
\]  
Since \(\|\hat{\Lambda}(\tau_j) - \Lambda(\tau_j)\| = O_p(n^{-3/10})\) and \(\Lambda^{-1}(\tau_j)\) is bounded:  
\[
\hat{\Lambda}^{-1}(\tau_j) - \Lambda^{-1}(\tau_j) = O_p(n^{-3/10}) + O_p(n^{-6/10}) = O_p(n^{-3/10}).
\]

\textit{Step 7: Synthesizing Results.}  
From Step 1:  
\[
\hat{D}(\tau_j \mid \mathbf{x}^*) - D(\tau_j \mid \mathbf{x}^*) = \mathbf{x}^{*^\top} \left[ (\hat{\Lambda}^{-1}(\tau_j) - \Lambda^{-1}(\tau_j)) \bar{\bx} + \Lambda^{-1}(\tau_j) (\bar{\bx} - \mathbb{E}[\bx]) \right].
\]  
\begin{itemize}
	\item First term: \(\mathbf{x}^{*^\top} (\hat{\Lambda}^{-1}(\tau_j) - \Lambda^{-1}(\tau_j)) \bar{\bx} = O_p(n^{-3/10})\), since \(\hat{\Lambda}^{-1}(\tau_j) - \Lambda^{-1}(\tau_j) = O_p(n^{-3/10})\) and \(\bar{\bx} = O_p(1)\). 
	\item Second term: \(\mathbf{x}^{*^\top} \Lambda^{-1}(\tau_j) (\bar{\bx} - \mathbb{E}[\bx]) = O_p(n^{-1/2})\), since \(\bar{\bx} - \mathbb{E}[\bx] = O_p(n^{-1/2})\).  
\end{itemize}
Total error:  
\[
\hat{D}(\tau_j \mid \mathbf{x}^*) - D(\tau_j \mid \mathbf{x}^*) = O_p(n^{-3/10}) + O_p(n^{-1/2}) = O_p(n^{-3/10}),
\]  
since \(-3/10 > -1/2\). 

Since the above derivations hold uniformly for all \(\tau_j \in [\tau_{\mathsf{L}}, \tau_{\mathsf{U}}]\), and given the compactness of the interval \([\tau_{\mathsf{L}}, \tau_{\mathsf{U}}]\), we conclude that:
\[
\max_{\tau_j \in [\tau_{\mathsf{L}}, \tau_{\mathsf{U}}]} \left| \hat{D}(\tau_j \mid \mathbf{x}^*) - D(\tau_j \mid \mathbf{x}^*) \right| = O_p\left( n^{-3/10} \right),
\]
thus completing the proof. \hfill $\square$

\subsubsection*{Proof of Theorem~\ref{thm: rate}}

\textit{Step 1: Error Decomposition.}  
First, we decompose the total error into two parts: 
\[
\hat{Q}_C(\tau \mid \mathbf{x}^*) - Q(\tau \mid \mathbf{x}^*) = \underbrace{\big( \hat{Q}_C(\tau \mid \mathbf{x}^*) - Q_C(\tau \mid \mathbf{x}^*) \big)}_{\text{Estimation Error}} ~ + \underbrace{\big( Q_C(\tau \mid \mathbf{x}^*) - Q(\tau \mid \mathbf{x}^*) \big)}_{\text{Interpolation Error}}.
\]  
where \( Q_C(\tau \mid \mathbf{x}^*) \) is the cubic Hermite interpolation using true values \( Q(\tau_j \mid \mathbf{x}^*) \) and \( D(\tau_j \mid \mathbf{x}^*) \). The first term is the error due to estimation, while the second term is the interpolation error.

\textit{Step 2: Bounding Interpolation Error.}  
By Lemma \ref{lemma:interpolation_error}, for the interpolation error, we have:  
\[
\sup_{\tau \in [\tau_{\mathsf{L}} , \tau_{\mathsf{U}}]} \left| Q_C(\tau \mid \mathbf{x}^*) - Q(\tau \mid \mathbf{x}^*) \right| = O\left( \frac{1}{m^4} \right).
\]  

\textit{Step 3: Bounding Estimation Error.} 
For any $j \in \{1, \ldots, m-1\}$ such that $[\tau_j, \tau_{j+1}] \cap [\tau_{\mathsf{L}} + \frac{1}{m}, \tau_{\mathsf{U}} - \frac{1}{m}] \neq \emptyset$, let $\tau \in [\tau_j, \tau_{j+1}] \cap [\tau_{\mathsf{L}} + \frac{1}{m}, \tau_{\mathsf{U}} - \frac{1}{m}]$. The cubic Hermite interpolants using estimated values and true values are:  
\[
\hat{Q}_C(\tau \mid \mathbf{x}^*) = h_{00}(\xi) \hat{Q}_j(\tau \mid \mathbf{x}^*) + \frac{h_{10}(\xi)}{m} \hat{D}_j(\tau \mid \mathbf{x}^*) + h_{01}(\xi) \hat{Q}_{j+1}(\tau \mid \mathbf{x}^*) + \frac{h_{11}(\xi)}{m} \hat{D}_{j+1}(\tau \mid \mathbf{x}^*),
\]  
\[
Q_C(\tau \mid \mathbf{x}^*) = h_{00}(\xi) Q_j(\tau \mid \mathbf{x}^*) + \frac{h_{10}(\xi)}{m} D_j(\tau \mid \mathbf{x}^*) + h_{01}(\xi) Q_{j+1}(\tau \mid \mathbf{x}^*) + \frac{h_{11}(\xi)}{m} D_{j+1}(\tau \mid \mathbf{x}^*),
\]  
where \(\xi = m(\tau - \tau_j) \in [0, 1]\), and the Hermite basis functions are:
\( h_{00}(\xi) = 2\xi^3 - 3\xi^2 + 1 \),
\( h_{10}(\xi) = \xi^3 - 2\xi^2 + \xi \),
\( h_{01}(\xi) = -2\xi^3 + 3\xi^2 \),
\( h_{11}(\xi) = \xi^3 - \xi^2 \). 
Then, the estimation error becomes: 
\begin{eqnarray*}
	\lefteqn{
		\hat{Q}_C(\tau \mid \mathbf{x}^*) - Q_C(\tau \mid \mathbf{x}^*)  }\\
	&=&  \ h_{00}(\xi) \left( \hat{Q}_j(\tau \mid \mathbf{x}^*) - Q_j(\tau \mid \mathbf{x}^*) \right) 
	+ \frac{h_{10}(\xi)}{m} \left( \hat{D}_j(\tau \mid \mathbf{x}^*) - D_j(\tau \mid \mathbf{x}^*) \right) \\
	&+&
	h_{01}(\xi) \left( \hat{Q}_{j+1}(\tau \mid \mathbf{x}^*) - Q_{j+1}(\tau \mid \mathbf{x}^*) \right)+ \frac{h_{11}(\xi)}{m} \left( \hat{D}_{j+1}(\tau \mid \mathbf{x}^*) - D_{j+1}(\tau \mid \mathbf{x}^*) \right).
\end{eqnarray*}
Using bounds on the Hermite basis functions over \( \xi \in [0, 1] \)
(\(|h_{00}(\xi)| \leq 1,\) \( |h_{01}(\xi)| \leq 1\);
\(|h_{10}(\xi)|\leq \frac{4}{27}, |h_{11}(\xi)| \leq \frac{4}{27}\)),
we have the bound:
\begin{eqnarray*}
	\lefteqn{
		\left| \hat{Q}_C(\tau \mid \mathbf{x}^*) - Q_C(\tau \mid \mathbf{x}^*) \right| }\\
	&\leq&  \left| \hat{Q}(\tau_j \mid \mathbf{x}^*) - Q(\tau_j \mid \mathbf{x}^*) \right| + \frac{4}{27m} \left| \hat{D}(\tau_j \mid \mathbf{x}^*) - D(\tau_j \mid \mathbf{x}^*) \right| \\
	&+&
	\left| \hat{Q}(\tau_{j+1} \mid \mathbf{x}^*) - Q(\tau_{j+1} \mid \mathbf{x}^*) \right| + \frac{4}{27m} \left| \hat{D}(\tau_{j+1} \mid \mathbf{x}^*) - D(\tau_{j+1} \mid \mathbf{x}^*) \right|.
\end{eqnarray*}
Taking the supremum over \( \tau \in [\tau_{\mathsf{L}}+\frac{1}{m}, \tau_{\mathsf{U}}-\frac{1}{m}] \) across all subintervals:
\begin{eqnarray*}
	\lefteqn{
		\sup_{\tau \in [\tau_{\mathsf{L}}+\frac{1}{m}, \tau_{\mathsf{U}}-\frac{1}{m}]} \left| \hat{Q}_C(\tau \mid \mathbf{x}^*) - Q_C(\tau \mid \mathbf{x}^*) \right| }\\
	&\leq&  2 \max_{\tau_j \in [\tau_{\mathsf{L}}, \tau_{\mathsf{U}}]} \left| \hat{Q}(\tau_j \mid \mathbf{x}^*) - Q(\tau_j \mid \mathbf{x}^*) \right| + \frac{8}{27m} \max_{\tau_j \in [\tau_{\mathsf{L}}, \tau_{\mathsf{U}}]} \left| \hat{D}(\tau_j \mid \mathbf{x}^*) - D(\tau_j \mid \mathbf{x}^*) \right|.
\end{eqnarray*}
By Lemma~\ref{lemma:qr_estimation_error} and \ref{lemma:gradient_estimation_error}, we have:  
\[
\max_{\tau_j \in [\tau_{\mathsf{L}}, \tau_{\mathsf{U}}]} \left| \hat{Q}(\tau_j \mid \mathbf{x}^*) - Q(\tau_j \mid \mathbf{x}^*) \right| = O_{\mathbb{P}}\left( \frac{1}{\sqrt{n}} \right),
\] 
\[
 \max_{\tau_j \in [\tau_{\mathsf{L}}, \tau_{\mathsf{U}}]} \left| \hat{D}(\tau_j \mid \mathbf{x}^*) - D(\tau_j \mid \mathbf{x}^*) \right| = O_{\mathbb{P}}\left( n^{-3/10} \right).
\]  
Thus:  
\[
\sup_{\tau \in [\tau_{\mathsf{L}}+\frac{1}{m}, \tau_{\mathsf{U}}-\frac{1}{m}]} \left| \hat{Q}_C(\tau \mid \mathbf{x}^*) - Q_C(\tau \mid \mathbf{x}^*) \right| = O_{\mathbb{P}}\left( \frac{1}{\sqrt{n}} \right) + O_{\mathbb{P}}\left( \frac{n^{-3/10}}{m} \right).
\]  

\textit{Step 4: Synthesizing Results.}  
Combining Steps 1–3:  
\begin{eqnarray*}
	\lefteqn{
		\sup_{\tau \in [\tau_{\mathsf{L}}+\frac{1}{m}, \tau_{\mathsf{U}}-\frac{1}{m}]} \left| \hat{Q}_C(\tau \mid \mathbf{x}^*) - Q(\tau \mid \mathbf{x}^*) \right| }\\
	&\leq&  \sup_{\tau \in [\tau_{\mathsf{L}}+\frac{1}{m}, \tau_{\mathsf{U}}-\frac{1}{m}]} \left| \hat{Q}_C(\tau \mid \mathbf{x}^*) - Q_C(\tau \mid \mathbf{x}^*) \right| +\sup_{\tau \in [\tau_{\mathsf{L}}+\frac{1}{m}, \tau_{\mathsf{U}}-\frac{1}{m}]} \left| Q_C(\tau \mid \mathbf{x}^*) - Q(\tau \mid \mathbf{x}^*) \right| \\
	& = & O_{\mathbb{P}}\left( \frac{1}{\sqrt{n}} \right) + O_{\mathbb{P}}\left( \frac{n^{-3/10}}{m} \right) + O\left( \frac{1}{m^4} \right),
\end{eqnarray*}
which completes the proof. \hfill $\square$

\section{Inventory Management}\label{app: inventory}
We consider the $(s,S)$ inventory management benchmark from the SimOpt library \citep{Eckman_SimOpt}. This test problem has been frequently adopted in the simulation optimization literature; see, e.g., \citet{li2025additive}.  The system evolves in discrete time. Under an $(s,S)$ replenishment rule, the inventory position is reviewed each period; if it falls below the reorder point $s$, an order is placed to bring the position up to the order-up-to level $S$. Customer demand in each period is exponentially distributed with mean $\mu$, and replenishment lead times follow a Poisson distribution with rate $\theta$. Costs accrue through three components: holding cost at rate $h$, a fixed ordering cost $f$ whenever an order is placed, and a per-unit purchasing cost $c$.

In our experiments, we view this problem as a stochastic simulator that takes $(s,S,\mu)$ as input and returns the corresponding average cost (from which we compute the target estimand). We fix the initial inventory to $1000$ and simulate a horizon of $1000$ periods. Lead times are Poisson with mean $\theta=6$, and we set $h=0.5$, $f=36$, and $c=1$. 

For dataset construction, we consider the domain \(s\in[270,340]\), \(S\in[380,450]\), and \(\mu\in[310,340]\). We uniformly generate \(n=10^4\) integer-valued covariate points from this domain and run the simulator once at each point to obtain the corresponding output \(y\). To evaluate KS and WD, we fix a covariate value \(\bx=\bx^*\) and run the simulator \(K=10^5\) times to construct a reference set. We then generate \(K=10^5\) observations from each generative model at the same \(\bx^*\), and compute KS and WD by comparing the model outputs against the reference set. To evaluate confidence-interval coverage, we establish the ground truth estimands empirically by calculating them from $10^5$ simulator observations.

\section{Implementation details}\label{app:detailed exp}

For all the experiments in Section \ref{subsec: synthetic exp} and \ref{subsec: practical data}, we generate a training set for each test distribution $\{(\bx_i,y_i)\}_{i=1}^n$ with $n=10^4$ to fit generative models, and generate $K=10^5$ observations from every model to evaluate performance. For the synthetic datasets, the covariate $ \bx^* = (1,4,-1,3)^\top$; for the inventory management dataset, the covariate \(\bx^* = (1,320,420,330)^\top\). All results are presented as mean values with standard errors derived from $N=100$ independent experimental replications. To ensure a fair comparison regarding training time, we maintain consistent CPU utilization across models. Specifically, for the experiments in Figure \ref{fig:E-QRGMM and QRGMM} and \ref{fig:n100000 E-QRGMM and QRGMM}, QRGMM and E-QRGMM are executed directly with identical CPU usage. In contrast, for the comparisons in Table \ref{tab:performance} and \ref{tab:performance inventory}, we employ parallel computing for E-QRGMM to scale its CPU utilization to match that of the deep generative models. In addition to the training data and testing methodology, we proceed to detail the individual settings of each model.

\paragraph{E-QRGMM} The general parameter settings are \(c = 2\), \(\tau_\mathsf{L} = 0.1\), \(\tau_\mathsf{U} = 0.9\), \(\delta_n = 0.1\), \(\alpha=0.1\), \(B=100\). For three synthetic datasets, we use simple linear quantile regression 
\(\hat{Q}(\tau \mid \bx) = \bx^\top \hat{\bbeta}(\tau)\) with \(m = \sqrt{n} = 100\). For the inventory management dataset, we employ quantile regression 
\(\hat{Q}(\tau \mid \bx) = \bb^\top(\bx)\hat{\bbeta}(\tau)\) with \(m = 300\), 
where \(\bx = (1, s, S,\mu)^\top\) and \(\bb(\bx)\) denotes a vector of  basis functions:
\[
\bb(\bx) = (1, s+S, S-s, 1/(S-s), \mu)^\top.
\]

 In experiments shown in the left panel of Figure~\ref{fig:E-QRGMM and QRGMM} and~\ref{fig:n100000 E-QRGMM and QRGMM}, we set the values of \(m\) in \(\mathcal{T}(m)\) to 
\([12, 30, 50, 70,\)
\(100,130, 160, 200, 250, 300, 400, 500, 600, 700]\).
\paragraph{QRGMM} The grid points in QRGMM form a uniform \(1/m\)-spaced grid: \(\{ j/m \}_{j=1}^{m-1}\), and the general default setting for \(m\) is also taken to be \(100\). To compare with E-QRGMM using the same number of grid points, we set the values of \(m\) for QRGMM to
\([9, 13, 17, 23, 31, 37, 45,\) \(53, 65, 75, 97, 119, 141, 161]\) 
in the experiment shown in the left panel of Figure~\ref{fig:E-QRGMM and QRGMM} and~\ref{fig:n100000 E-QRGMM and QRGMM}.

\paragraph{GAN} 
Following \citet{liang2024generative}, the generator \( G : \mathbb{R}^{p + d} \to \mathbb{R} \) is designed to transform a simple distribution into the target conditional distribution, and the discriminator \( D : \mathbb{R}^{p + 1} \to (0,1) \) is used to assess the quality of the generated observations. The loss is defined as:
\[
\min_{G} \max_{D} \, 
\mathbb{E}_{(Y,\bx) \sim p(Y\mid \bx)} [\log D(Y, \bx)] 
+ 
\mathbb{E}_{\bx, z \sim \mathcal{N}(0, I_d)} [\log(1 - D(G(\bx, z), \bx))].
\]

To generate observations, we draw latent variates \( z_k \sim \mathcal{N}(0, I_d), k=1,\ldots,K\), combine them with the condition \(\bx= \bx^*\), and generate predictions via \( y_k = G(\bx^*, z_k) \). Both the generator \(G\) and discriminator \(D\) are implemented as multilayer perceptrons (MLPs).  We set \(\texttt{latent\_dim} = d = 1\), \(\texttt{hidden\_dim} = 128\), and the number of hidden layers \(\texttt{num\_layer} = 3\) for generator with \texttt{ReLU} activation at the output.  The discriminator shares the same hidden dimension and number of layers as the generator, but applies a \texttt{sigmoid} activation at the output. Both the networks use \texttt{ReLU} activation in hidden layers.

\paragraph{DDIM} Our implementation utilizes the default configuration in the denoising diffusion implicit model (DDIM)\citep{song2020denoising}. The noise estimation network is implemented as a multilayer perceptron (MLP), defined as
\(
\varepsilon_\theta(\bx, z, t): \mathbb{R}^{p + d + 1} \rightarrow \mathbb{R},\)
where \(\bx \in \mathbb{R}^p\) is the covariate, \(z \in \mathbb{R}^d\) is the noisy latent input, and \(t\) is the normalized time step. The MLP shares the same architecture and hyperparameters as the generator used in the GAN model. For data generation, we use a 10-step discretization of the reverse process.

\paragraph{RectFlow}
Our implementation builds upon the frameworks proposed by \citet{liu2022flow} and \citet{liang2024generative}. The transformation ODE from a smooth distribution \(q(Y)\) (e.g., Gaussian) to covariate-dependent target distribution \(p(Y \mid \bx)\) can be obtained by learning vector field
\[
u(Y, t, \bx) = \mathbb{E}_{Y_0 \sim q(Y),\, Y_1 \sim p(Y \mid \bx)} \left[ Y_1 - Y_0 \mid Y_t = Y \right],
\]
where \( Y_t = (1 - t)Y_0 + t Y_1 \), \(Y_0\sim q(Y)\) is sampled from a fixed prior distribution (e.g., Gaussian), \(Y_1\sim p(Y\mid \bx)\) is from the training dataset. Given data from \(p(Y \mid \bx)\), we train a neural network \(v_\theta(Y, t, \bx)\) to approximate the vector field.
To ensure robust performance across diverse input covariates, the loss for training is as follows (see \citet{liang2024generative} for details):

\[
\mathcal{L}(v_\theta) = \mathbb{E}_{\bx} \left[ \mathcal{L}_{\bx}(v_\theta) \right] 
= \int_0^1 \mathbb{E}_{Y_0 \sim q(Y),\, Y_1 \sim p(Y \mid \bx)} 
\left[ \left\| v_\theta(Y_t, t, \bx) - (Y_1 - Y_0) \right\|^2 \right] dt.
\]

The neural network $v_\theta$ shares the same MLP architecture and hyperparameters as the generator used in the GAN model. Given a new input \(\bx^*\), we approximately generate observations from \(p(Y \mid \bx^*)\) by solving the ODE 
\(Y_1 = Y_0 + \int_0^1 v_\theta(Y_t, t, \bx^*) \, dt\), where \(Y_t = (1 - t)Y_0 + t Y_1\), using the Euler method with 10 discrete steps.

All deep generative models are trained on a dataset of size \(n = 10000\) for 1000 iterations with a batch size of 1000, using the Adam optimizer with a learning rate of \(10^{-3}\) and a weight decay of \(10^{-6}\). Before training, both the input covariates and the simulation outputs are standardized, and generated outputs are transformed back to their original scale for evaluation. This setup consistently yields accurate and stable results.

\end{document}